\documentclass[%
 aip,
 amsmath,amssymb,
 reprint,%
]{revtex4-2}

\usepackage{graphicx}
\usepackage{dcolumn}
\usepackage{bm}

\usepackage[utf8]{inputenc}
\usepackage[T1]{fontenc}
\usepackage{mathptmx}
\usepackage{etoolbox}

\usepackage[breaklinks]{hyperref}  
\hypersetup{
  colorlinks,
  citecolor=blue,
  linkcolor=blue,
  urlcolor=blue}
\usepackage[caption=false,position=top,margin=12pt,justification=raggedright,singlelinecheck=false,subrefformat=parens,labelformat=parens]{subfig}
\usepackage{subfig}
\makeatletter
\def\@email#1#2{%
 \endgroup
 \patchcmd{\titleblock@produce}
  {\frontmatter@RRAPformat}
  {\frontmatter@RRAPformat{\produce@RRAP{*#1\href{mailto:#2}{#2}}}\frontmatter@RRAPformat}
  {}{}
}%
\makeatother
\begin{document}

\preprint{AIP/123-QED}

\title[Learning Spatiotemporal Chaos Using Next-Generation Reservoir Computing]{Learning Spatiotemporal Chaos Using Next-Generation Reservoir Computing}
\author{Wendson A. S. Barbosa}
\email{desabarbosa.1@osu.edu.}
\affiliation{ 
Department of Physics, The Ohio State University, 191 W. Woodruff Ave., Columbus, OH 43210, USA 
}%

\author{Daniel J. Gauthier}%
 \email{gauthier.51@osu.edu.}
\affiliation{ 
Department of Physics, The Ohio State University, 191 W. Woodruff Ave., Columbus, OH 43210, USA 
}%
\affiliation{ResCon Technologies, LLC, PO Box 21229, Columbus, OH 43221, USA.}


\begin{abstract}
Forecasting the behavior of high-dimensional dynamical systems using machine learning requires efficient methods to learn the underlying physical model. We demonstrate spatiotemporal chaos prediction using a machine learning architecture that, when combined with a next-generation reservoir computer, displays state-of-the-art performance with a computational time $10^3-10^4$ times faster for training process and training data set $\sim 10^2$ times smaller than other machine learning algorithms. We also take advantage of the translational symmetry of the model to further reduce the computational cost and training data, each by a factor of $\sim$10. 

\end{abstract}

\maketitle

\begin{quotation}

Modeling and predicting high-dimensional dynamical systems, such as spatiotemporal chaotic systems, continues to be a physics grand challenge and require efficient methods to leverage computational resources and efficiently process large amounts of data. 
In this work, we implement a highly efficient machine learning (ML) parallel scheme for spatiotemporal forecasting where each model unit predicts a single spatial location. This reduces the number of trainable parameters to the minimum number possible, thus speeding up the algorithm and reducing the data set size needed for training. 
Moreover, when combined with 
next-generation reservoir computers (NG-RCs), our approach presents state-of-the-art performance with a computational cost and training data dramatically reduced in comparison to other machine learning approaches.
We also show that the computational cost and training data set size can be further reduced when the system display translational symmetry, which is commonly present in spatiotemporal systems with cyclic boundary conditions. Although many real systems do not have such symmetry, our results highlight the importance of symmetry addressing when it is present in the system.  

\end{quotation}

\section{\label{sec:Intro}Introduction}

Many nonlinear systems display temporal dynamics that depend on spatial location, such as the heart,\cite{winfree1987time} optical devices,\cite{ILLING2007bib} and fluid flow. \cite{holmes_lumley_berkooz_1996} These systems may displays spatiotemporal chaos, which has finite spatiotemporal correlations, a loss of long-term predictability,\cite{Lorenz1996} and the appearance of coherent structures.\cite{holmes_lumley_berkooz_1996} Also, these systems often display multi-scale behavior, where information and energy flow across scales.  Modelling spatiotemporal chaos is difficult for these reasons and continues to be a physics grand challenge.

One approach to this problem is to use machine learning, which may speed up prediction by learning only variables of interest, such as the macroscale behavior,\cite{Wilks2005,Chattopadhyay2019,Pyle2021} or improve the prediction accuracy by fusing model predictions and experimental observations.\cite{Abarbanel2018,Wikner2020,Fan2020,Wikner2021,Arcomano2022,Chattopadhyay2022}  Some researchers use ML algorithms, such as deep learning,\cite{vlachas2019,Chattopadhyay2019} time embedding techniques\cite{Parlitz2000,Orstavik1998,NG-RC} or sparse system identifiers,\cite{Lai2021} to learn the underlying ordinary or partial differential equations, but this subsequently requires precise numerical methods for model integration.  Another approach is to learn the system \textit{flow}, which allows for one-step-ahead prediction using a coarser spatiotemporal grid, likely leading to faster prediction. The next-generation reservoir computer (NG-RC), for example, excels at this task,\cite{NG-RC} and is mathematically equivalent to a traditional reservoir computer (RC) but has an optimal form.\cite{Bollt}

In typical ML approaches, the spatial variable is discretized at $L$ points with step size $\delta L$, assumed one-dimensional for exposition simplicity, and time is discretized with step size $\delta t$.  During supervised training, blocks of data with $N_{in}$ spatial points and $k$ time steps are fed into the artificial neural network (ANN) used to predict the behavior at $N_{out}$ locations at one or more temporal steps.  Often, $N_{in}=N_{out}=L$ and $k \delta t$ is longer than the correlation time,\cite{Pathak2018Hybrid,Chattopadhyay2019} which is problematic because it causes the model to focus on unrelated observations.  Also, the ANN is large, which increases the number of trainable parameters and hence increases the required computer resources and training data set size.  Recently, a parallel reservoir computing scheme\cite{Pathak2018} was introduced with $N_{out} < N_{in} < L$ to reduce the computational cost.

\begin{figure}[!ht]
\centering
	\subfloat{%
	\includegraphics[width=\linewidth]{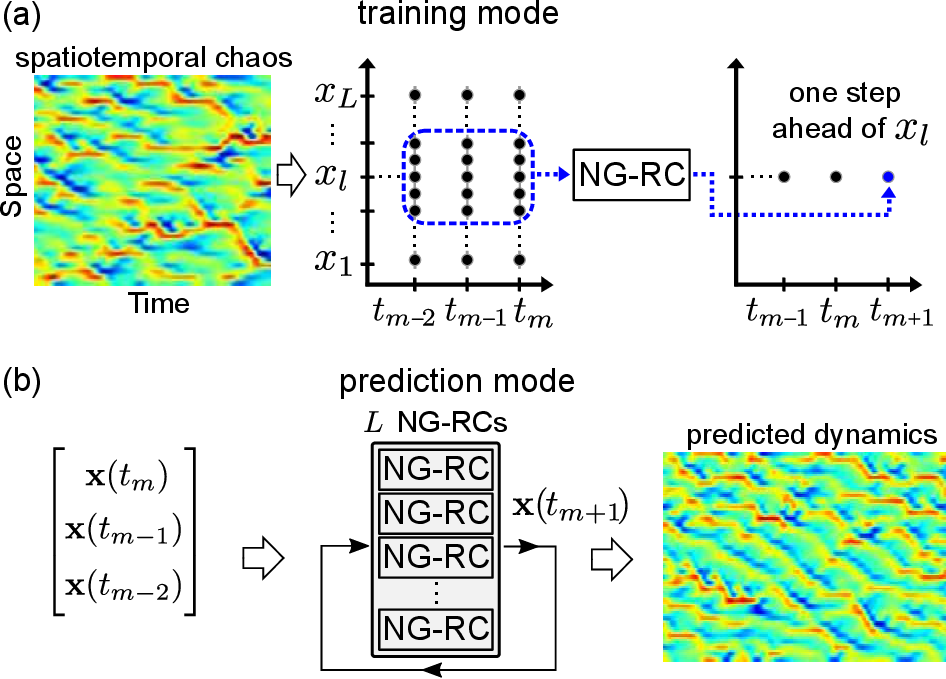}%
	\label{fig:ML_scheme_a}%
	}
	\begin{minipage}[t]{0\textwidth}
	\subfloat{%
	\label{fig:ML_scheme_b}%
	}
    \end{minipage}
\caption{Learning and predicting spatiotemporal chaos using our parallel scheme.  (a) \textbf{Learning Mode:} The NG-RC is trained to predict the next time step $t_{m+1}$ at the $l^{th}$ spatial location using training data from the current $t_m$ and previous steps $t_{m-1}$ and $t_{m-2}$ ($k=3$), and the $N_{in}=5$ neighbors. (b) \textbf{Prediction mode:} Autonomous operation of $L$ parallel NG-RCs where the output feeds the input to predict the next step of all spatial locations ${\bf{x}}=[x_1,x_2,\dots,x_L]$.}
\label{fig:ML_scheme}
\end{figure}

Here, our primary contribution is to demonstrate a new algorithm for learning spatiotemporal systems. It makes predictions of the temporal dynamics of the system at a single spatial location based on data drawn from a small spatiotemporal neighborhood of the point.  Prediction over all spatial points is realized using parallel machines. Quantitatively, we take $N_{in} < L$ and $N_{out}=1$, where $N_{in} \delta L$ is less than or comparable to the spatial correlation length, and take $k$ so that $k \delta t$ is less than the correlation time.  Hence, we use $L$ parallel ANNs for one-step-ahead prediction.  We apply our method to a heuristic atmospheric weather model using the NG-RC \cite{NG-RC} as the core learning machine, which reduces the computational complexity and the data set size required for training while displaying state-of-the-art accuracy. Accounting for the translational symmetry of this model further reduces the training computational time and data. Figure \ref{fig:ML_scheme} illustrates our scheme.

We highlight that addressing symmetries has proven to be important for improving other ML approaches.\cite{Lu2017,Herteux2020,BarbosaPRE2021,Favoni2022,OberlackPRL2022}  Furthermore, ML algorithms can reveal hidden symmetries, such as translational invariance present in a simple 1D uniform motion or black hole dynamics,\cite{Liu2022} even if their presence is not obvious.

The rest of the paper is organized as follows. In Sec. \ref{sec:L96}, we formally describe the high-dimensional dynamical system used as the learning system to train our ML approach for spatiotemporal chaos prediction. 
In Sec. \ref{sec:ParallelScheme}, we describe our parallel scheme of ML models where each model unit predicts a single discretized spatial location represented by a system variable.
In Sec. \ref{sec:NGRC} we introduce the theoretical background of the NG-RC,
followed by brief descriptions of the training procedure and prediction mode. 
Section \ref{sec:Results} is dedicated to the results. Here, we use our parallel scheme of NG-RCs to forecast high-dimensional chaos from the model equations introduced in Sec. \ref{sec:L96}. We compare the cases where the parallel NG-RCs are trained independently to the case where a translational symmetry is taken into account to improve the performance. Finally, we apply our approach to a lower dimensional case of Lorenz96 model system and to an even lower dimensional case where fine-scales variables are not present. Lastly, we present a discussion comparing our results to other works and our conclusions in Sec.  \ref{sec:Discussion_Conclusion}. We also include details on Ridge regression parameter optimization and computational complexity in the appendix.

\section{\label{sec:L96}Extended Lorenz96 model}

We demonstrate our approach using numerically generated data from a heuristic atmospheric weather model introduced by Lorenz \cite{Lorenz1996,Lorenz2005} and extended by Thornes \textit{et al.}. \cite{Thornes2017} It has an unspecified macroscopic scalar variable $x_l$ on a discrete grid (position $l$) representing the observations around a latitude circle. To represent some convective-scale quantity across spatiotemporal scales, this variable is driven by a finer-scale variable $y_{j,l}$, which is coupled to the macroscopic variable as well as the finest scale variable $z_{i,j,l}$ representing, for example, individual clouds in the atmosphere.

The model is described by a set of coupled differential equations given by
 \begin{eqnarray}
 \dot{x}_l&=&x_{l-1}(x_{l+1}-x_{l-2}) -x_{l}+F- \frac{hc}{b}S_{y_l},\nonumber \\  
 \dot{y}_{j,l}&=&-cby_{j+1,l}(y_{j+2,l}-y_{j-1,l}) -cy_{j,l}+\frac{hc}{b}x_l -  \frac{he}{d}S_{z_{j,l}}, \nonumber \\ 
 \dot{z}_{i,j,l}&=&edz_{i-1,j,l}(z_{i+1,j,l}-z_{i-2,j,l}) - gez_{i,j,l}+ \frac{he}{d}y_{j,l}, 
\label{eq:L96}
\end{eqnarray}
where the indices $l=1,\ldots,L$, $j=1,\ldots,J$  and  $i=1,\ldots,I$  are $\mathrm{mod}(l,L)$, $\mathrm{mod}(j,J)$, and $\mathrm{mod}(i,I)$, respectively, to represent cyclic boundary conditions. The terms $S_{y_l}=\sum_{j=1}^Jy_{j,l}$ and $S_{z_{j,l}}=\sum_{i=1}^Iz_{i,j,l}$ represent the couplings between the different spatiotemporal scales.  

Here, $F=20$ is a spatially homogeneous, large-scale forcing term, $h=1$ is the coupling strength between the different spatial scales and the parameters $b=c=d=e=g=10$ set the magnitude and time scale of the fast variables. With these parameters, there is a factor of 100 difference in spatiotemporal scales from the finest ($z$) to the coarsest ($x$) scale. For future reference, we specify time in model time units (MTU), where 1 MTU corresponds approximately to 5 atmospheric days.\cite{Lorenz1996} We take $L=36$ to set the number of coherent structures appropriate for the Earth's weather and $J=10$.\cite{Lorenz1996} For the fastest variable, we take $I=J$ following previous studies.\cite{Chattopadhyay2019,Pyle2021} There are $L*J=360$ fine-scale and $L*J*I=3,600$ finest-scale variables and hence there are $L[1+J(1+I)]=3,996$ total variables.

We focus on learning and predicting only the slow macroscopic variables $x_l$ without observing the finer-scale dynamics.\cite{Chattopadhyay2019,ChattopadhyayTransfLearning2020,Pyle2021}  Because of the fast time scale of $y_{l,j}$ and $z_{i,j,l}$ in comparison to $x_l$, $S_{y_l}$ acts as a noise-like term in driving $x_l$.  It is known that many ML algorithms, including an NG-RC, \cite{NG-RC} can learn in the presence of large noise and hence we expect that we can make accurate predictions as demonstrated below.

\section{\label{sec:ParallelScheme} The parallel ML scheme}

Our goal is to learn the one-step-ahead dynamics at a single location $x_l$ based on using $N_{in} = (2 N_{nn}+1)$ spatial points and $k$ temporal points, illustrated by the dashed boundary shown in the middle panel of Fig.~\ref{fig:ML_scheme_a}, shown for $N_{nn}=2$, $N_{in}=5$, and $k=3$, values we use in the results section below. 
We seek an ML model that predicts $x_l$ at time step $t_{m+1}$ based on this input data. Thus, there are $L$ independent ML models to predict the dynamics at all spatiotemporal points.  Our scheme can work with a variety of ML algorithms but we use an NG-RC because of its proven ability to make accurate predictions with limited data and low computational resources.\cite{NG-RC}

\section{\label{sec:NGRC}The NG-RC}

We operate the NG-RC in two modes shown in Fig.~\ref{fig:ML_scheme}: training and forecasting.  In both cases, we create a feature vector 
 \begin{eqnarray}
{\mathcal O}_{l,total}(t_m)= c \bigoplus {\mathcal O}_{l,lin} \bigoplus {\mathcal O}_{l,nonlin}
\label{eq:Ototal}
\end{eqnarray}
composed of linear and nonlinear parts, respectively, where $c$ is a constant and $\bigoplus$ is the concatenation operator. The linear part is formed by the current and previous $k-1$ values of the variable $x_l$ and its $N_{nn}$ nearest neighbors of each side of this location.  Hence, the dimension of ${\mathcal O}_{l,nonlin}$ is $d_{lin}=kN_{in}=15$ for our choice of parameters.  

We take the nonlinear part to be the unique second-order monomials of ${\mathcal O}_{l,lin}$, which is appropriate for this problem because Eqs.~\ref{eq:L96} contains only quadratic nonlinear terms.  For the unique quadradic monomials, the dimension of ${\mathcal O}_{l,nonlin}$ is $d_{nonlin}=d_{lin}(d_{lin}+1)/2=120$ so that the total feature vector ${\mathcal O}_{l,total}$ has $d_{total}=1+d_{lin}+d_{nonlin}=136$ components, which is also equal to the number of trainable parameters for each NG-RC.  Minimizing $d_{total}$ is one important metric for reducing the computational resources during training.  As an aside, we mention that other nonlinear functions can be used in the NG-RC, such as higher-order polynomials or radial basis functions, but are not needed here.

During training, data from the solution to Eqs.~\ref{eq:L96} with $N_{in}$ spatial points and $M$ temporal points ($t_{train} = M \delta t$ training time) is fed into each NG-RC ($L$ total) in an open-loop manner as illustrated in Fig.~\ref{fig:ML_scheme_a}.  Here, the goal is to have the one-step-ahead prediction $\overline{x}_l(t_{m+1})$ of the NG-RC equal to the model prediction $x_l(t_{m+1})$, 
where $t_{m+1}=t_m+\delta t$.  That is, we seek a solution to 
 \begin{eqnarray}
\overline{x}_l(t_{m+1}) = \mathrm{\mathbf{W}}_l {\mathcal O}_{l,total}(t_m)
\label{eq:solution}
\end{eqnarray}
that minimizes $||\overline{x}_l-x_l||^2 + \alpha ||\textrm{\textbf{W}}_l||^2$ over all $M$ temporal points. Here, $\textrm{\textbf{W}}_l$ is found using regularized regression with regularization parameter $\alpha$ (see appendix \ref{app:optimization} for $\alpha$ optimization). For our parameters, $\textrm{\textbf{W}}_l$ is a $(1\times d_{total})=(1 \times 136)$ matrix.  For future reference, the computational complexity of training a single NG-RC scales as $Md_{total}^2$ and hence as $LMd_{total}^2$ for all $L$ NG-RCs.

After finding $\textrm{\textbf{W}}_l$, we switch to prediction mode, where the training data is no longer input.  The output of each NG-RC is sent to the input in a closed-loop manner and used to create ${\mathcal O}_{l,total}(t)$ for each NG-RC as shown in Fig.~\ref{fig:ML_scheme_b}.  The parallel NG-RCs now form an autonomous spatiotemporal dynamical system. Here, the `warm up' of the NG-RC requires $k$ previous time steps as the initial conditions, which is often available if we immediately switching to the forecasting mode from the training mode.

\section{\label{sec:Results}Results}

To test the accuracy and speed of our new algorithm, we generate spatiotemporal data by integrating Eqs.~\ref{eq:L96} using a fourth order Runge-Kutta method with fixed step size of 0.001 MTU, where we save data at steps of $\delta t=0.01$ MTU. We use initial conditions $x_1=F+0.01$, $x_{l\neq1}=F$ and $y_{j,l}=z_{i,j,l}=0$, integrate for 10 MTU to dissipate transients and discard this data. We integrate for an additional 11,000 MTU to generate the data from which we select an interval $t_{train}$ ($M=t_{train}/\delta t$) to use as training data set. After integration, we normalize the data to have zero mean and unit standard deviation. After training, we select $k$ consecutive time steps to warm up the NGRCs for the prediction mode and select the following testing time interval to generate the ground-truth test data (data never seen by the NG-RCs during training), an example of which is shown in Fig.~\ref{fig:SingleNGRC_a}.

\subsection{\label{sec:SingleNGRC} Single NG-RC}

\begin{figure}[t]
\centering
	\subfloat{%
	\includegraphics[width=\linewidth]{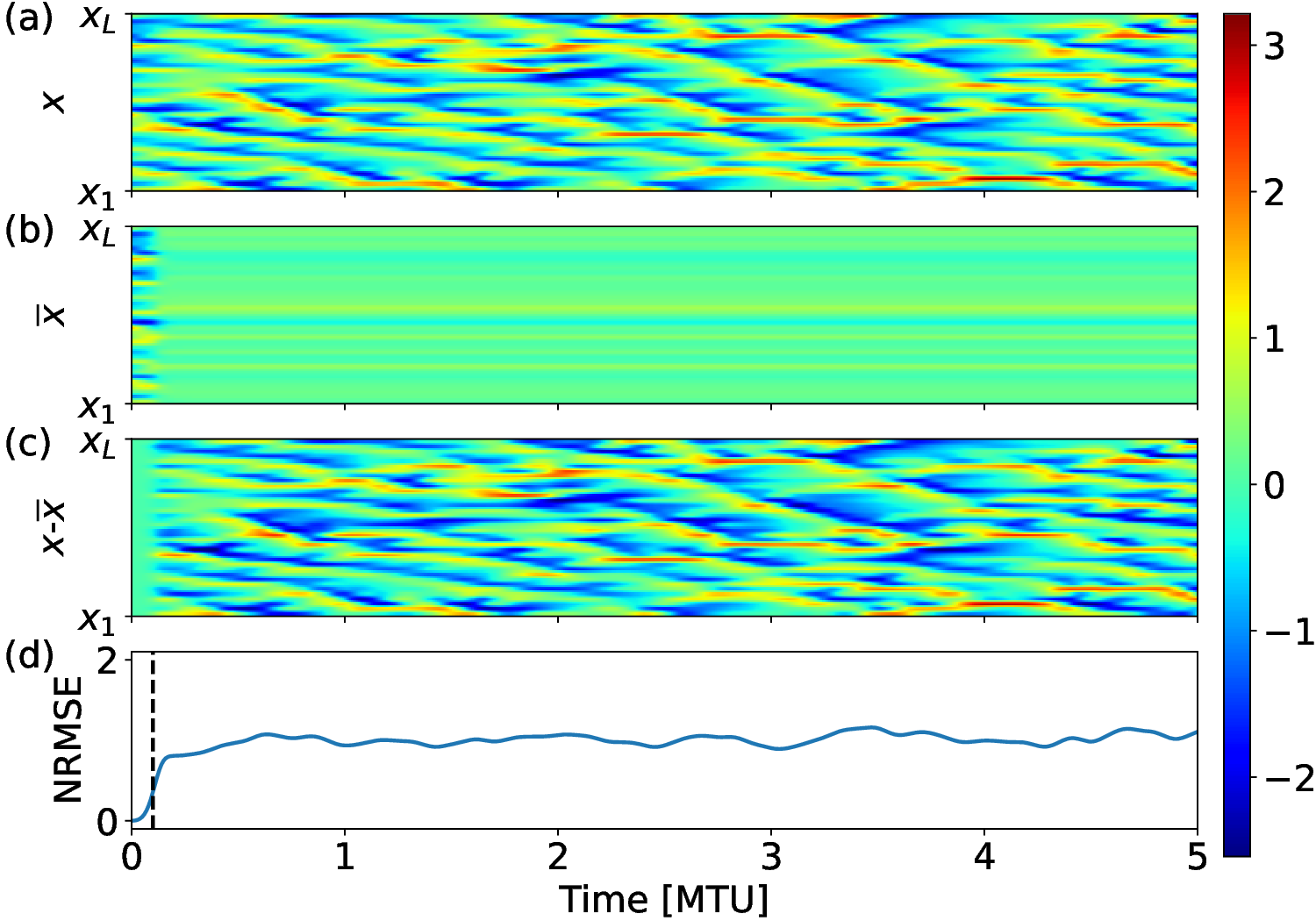}
	\label{fig:SingleNGRC_a}%
	}
	\begin{minipage}[t]{0\textwidth}
	\subfloat{%
	\label{fig:SingleNGRC_b}%
	}
   	 \end{minipage}
	\begin{minipage}[t]{0\textwidth}
	\subfloat{%
	\label{fig:SingleNGRC_c}%
	}
   	 \end{minipage}
	\begin{minipage}[t]{0\textwidth}
	\subfloat{%
	\label{fig:SingleNGRC_d}%
	}
   	 \end{minipage}
\caption{Spatiotemporal dynamics prediction with a non-parallel scheme with a single NG-RC. (a) Actual and (b) predicted dynamics for the extended Lorenz96 model.  (c) Difference and (d) NRMSE between actual and predicted dynamics. The vertical dashed line indicates the prediction horizon. Parameters: $k=3$, $N_{in}=N_{out}=L=36$ and $\alpha = 10^{-2}$.}
\label{fig:SingleNGRC}
\end{figure}

As a baseline, we first predict the extended Lorenz96 system using a non-parallel architecture formed by a single NG-RC. The model receives all $L$ variables as input and is trained to perform one-step ahead prediction at all spatial locations; \textit{i.e.}, $N_{in}=N_{out}=L$. Here, the dimension of ${\mathcal O}_{l,nonlin}$ is $d_{lin}=kN_{in}=108$ when $k=3$ as used here. Thus, the feature vector ${\mathcal O}_{l,total}$ has $d_{total}=5,995$ components. The large number of features causes the model to focus on unrelated observations outside of the spatial correlation length and the single NG-RC fails to predict the spatiotemporal dynamics after a short period. Figure~\ref{fig:SingleNGRC_b} shows the NG-RC-predicted spatiotemporal dynamics of the extended Lorenz96 model using a relatively small training data set size $t_{train}=10$ MTU ($M=1,000$). As we show in the next sections, the performance of this approach can be improved using a much larger large training data set, which helps the model learn the appropriated features and to filter out the uncorrelated ones. 

To quantify the prediction quality, we determine the normalized root-mean-square error over all spatial locations given by
\begin{equation}
    \mathrm{NRMSE}(t) =\sqrt{\frac{1}{L}\sum_{l=1}^L(x_l(t)-\overline{x}_l(t))^2}.
    \label{eq:NRMSE}
\end{equation}
The right-hand side of Eq.~\ref{eq:NRMSE} is already normalized because $x_l$ has unit standard deviation. 

As shown in Fig.~\ref{fig:SingleNGRC_d}, the NRMSE for the single NG-RC prediction increases from nearly zero, eventually reaching a saturated value. We define a prediction horizon as the time where NRMSE=0.3 (vertical dashed line), which is equal to 0.1 MTU for the single NG-RC prediction realization shown in Fig.~\ref{fig:SingleNGRC}. When averaged over 100 predictions for different initial conditions, the prediction horizon is equal to 0.01 $\pm$ 0.03  MTU, or approximately 0.05 $\pm$ 0.14 atmospheric days for the extended Lorenz96 model.

\subsection{\label{sec:LindependentNGRCs}$L$ independent NG-RCs}

\begin{figure}[t]
\centering
	\subfloat{%
	\includegraphics[width=\linewidth]{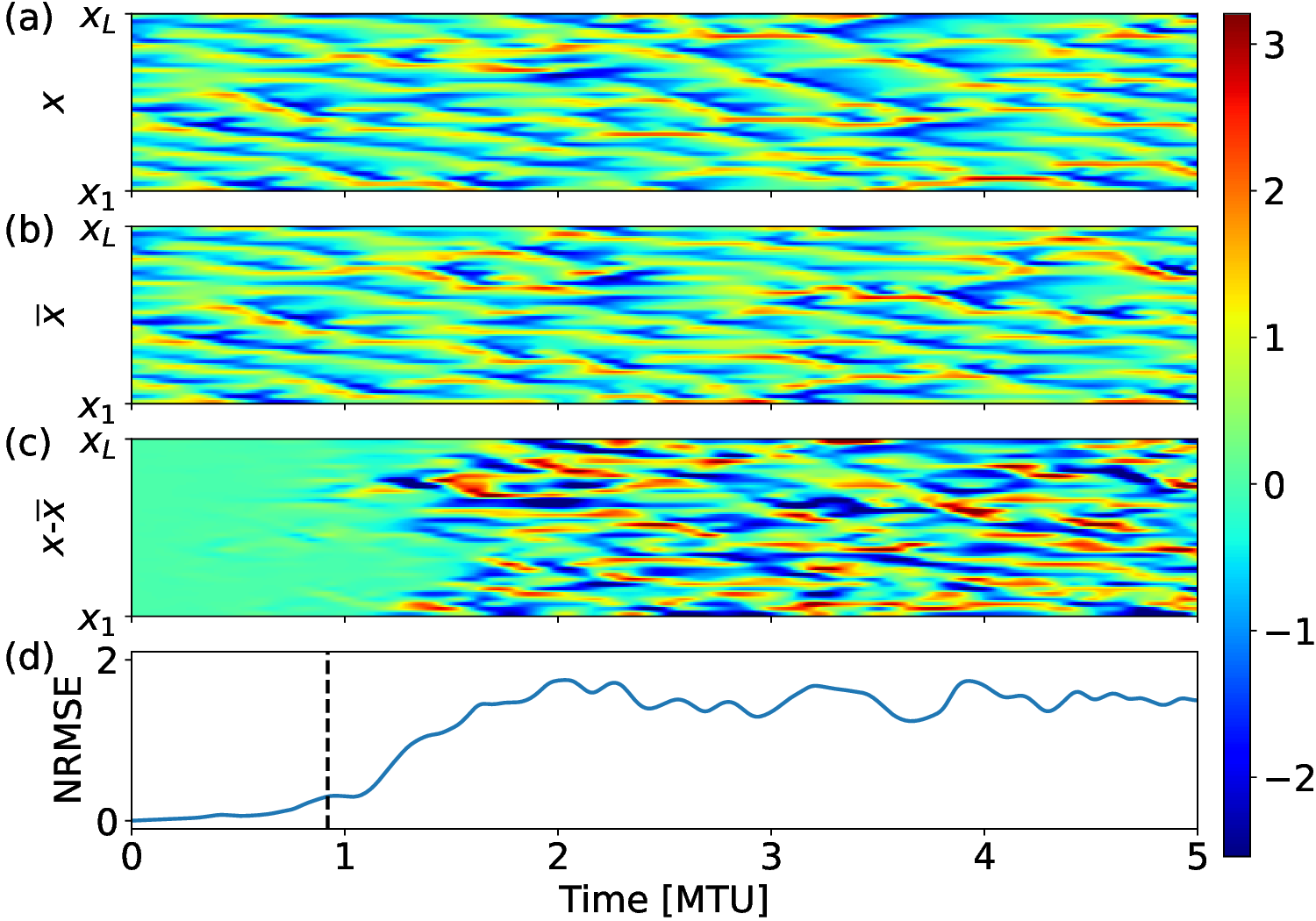}
	\label{fig:TrainingALLNG-RCs_a}%
	}
	\begin{minipage}[t]{0\textwidth}
	\subfloat{%
	\label{fig:TrainingALLNG-RCs_b}%
	}
   	 \end{minipage}
	\begin{minipage}[t]{0\textwidth}
	\subfloat{%
	\label{fig:TrainingALLNG-RCs_c}%
	}
   	 \end{minipage}
	\begin{minipage}[t]{0\textwidth}
	\subfloat{%
	\label{fig:TrainingALLNG-RCs_d}%
	}
   	 \end{minipage}
\caption{Spatiotemporal dynamics prediction with parallel NG-RCs using L independent $\mathbf{W}_l$'s. (a) Actual and (b) predicted dynamics for the extended Lorenz96 model.  (c) Difference and (d) NRMSE between actual and predicted dynamics. The vertical dashed line indicates the prediction horizon. Parameters: $k=3$, $N_{nn}=2$ and $\alpha = 10^{-2}$.}
\label{fig:TrainingALLNG-RCs}
\end{figure}

In this section, we implement our scheme that uses smaller NG-RCs operating in parallel. We train $L=36$ NG-RCs, each with $d_{total}=136$. In terms of computational complexity, training the $L$ parallel NG-RCs is around 50 times less expensive than training the single NG-RC used in section \ref{sec:SingleNGRC}, as the NG-RC computational complexity scales as $\mathcal{O}(Md_{total}^2)$ as we discuss later. Furthermore, we choose $N_{in}=5$ so that each parallel NG-RC focuses on data from nearby spatial locations within the spatial correlation distance. Figure~\ref{fig:TrainingALLNG-RCs_b} shows the NG-RC-predicted spatiotemporal dynamics of the extended Lorenz96 model using $t_{train}=10$ MTU ($M=1,000$), the same training data set as in the previous example. The difference between ground-truth and prediction are initially small (Fig.~\ref{fig:TrainingALLNG-RCs_c}), but eventually diverge because the chaotic nature of the system will amplify small difference between the two.  Even though long-term prediction is lost, the predicted behavior has coherent structures that are visually similar to the extended Lorenz96 model. Here, the prediction horizon is equal to 0.92 for the single prediction realization shown in Fig.~\ref{fig:TrainingALLNG-RCs}. When averaged over 100 predictions for different initial conditions, the prediction horizon is equal to 0.66 $\pm$ 0.15  MTU, or approximately 3.2 $\pm$ 0.7 atmospheric days for the extended Lorenz96 model.  We comment on the accuracy of our prediction in the discussion section below.

\subsection{\label{sec:LUsingSymmetry}Using translational symmetry}

\begin{figure}[t]
\centering
	\subfloat{%
	\includegraphics[width=\linewidth]{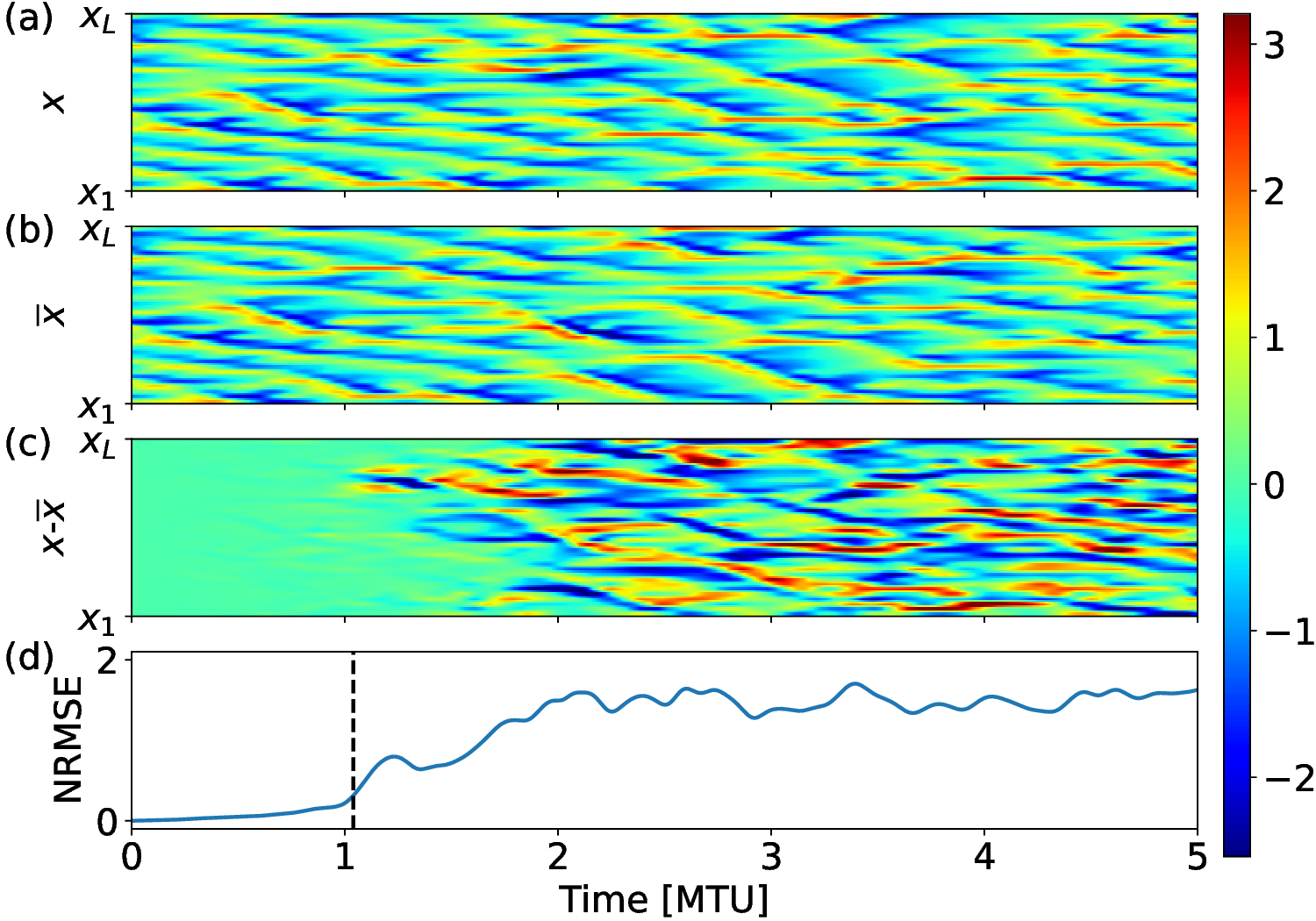}
	\label{fig:Training1NG-RC_a}%
	}
	\begin{minipage}[t]{0\textwidth}
	\subfloat{%
	\label{fig:Training1G-RC_b}%
	}
   	 \end{minipage}
	\begin{minipage}[t]{0\textwidth}
	\subfloat{%
	\label{fig:Training1NG-RC_c}%
	}
   	 \end{minipage}
	\begin{minipage}[t]{0\textwidth}
	\subfloat{%
	\label{fig:Training1NG-RC_d}%
	}
   	 \end{minipage}
\caption{Spatiotemporal dynamics prediction with parallel NG-RCs using a single $\mathbf{W}$ that respects translational symmetry. (a) Actual and (b) predicted dynamics for the extended Lorenz96 model. (c) Difference and (d) NRMSE between actual and predicted dynamics. The vertical dashed line indicates the prediction horizon. Parameters: $k=3$, $N_{nn}=2$ and $\alpha = 10^{-2}$.}
\label{fig:Training1NG-RC}
\end{figure}

We further decrease the training data set size by taking into account the translational invariance of the extended Lorenz96 model given in Eq.~\ref{eq:L96}, which arises from the cyclic boundary conditions and the spatially-independent model parameters. Because of this symmetry, $\mathbf{W}_l$ should be independent of $l$ in the asymptotic limit $t_{train} \rightarrow \infty$. 
To quantify the independence for the finite $t_{train}$ used here, we measure the similarity of the $\mathbf{W}_l$'s by defining a correlation coefficient 
\begin{equation}
    C =\frac{1}{L(L-1)}\sum_{l}^L \sum_{l'\neq l}^L\frac{ \mathbf{W}_l^T \boldsymbol{\cdot} \mathbf{W}_{l'}^T}{||\mathbf{W}_l||^2},
\end{equation}
where $T$ is the transpose operation, $\boldsymbol{\cdot}$ is the dot product operation, and $C=1$ ($C=0$) indicates (un)correlated matrices. For the case presented in Fig. \ref{fig:TrainingALLNG-RCs}, we find that it is equal to 0.96, indicating that the NG-RC does a reasonable job of discovering this symmetry even with short $t_{train}$.

We force translational symmetry by training a single $\mathbf{W}$ and use it for all spatial locations $l$.  Operationally, we concatenate all ${\mathcal O}_{l,total}$ to create a data structure that has dimension $LMd_{total}$ and hence the training computational complexity scales as $LMd_{total}^2$, the same as in the previous scheme. This procedure relies on the fact that each spatial location has identical behavior - in a statistical sense - and hence all the data is produced by the same underlying dynamical flow. Effectively, it increases the training time to $Lt_{train}$ for an observation time $t_{train}$ of the spatiotemporal dynamics. A similar approach has been used recently when using a traditional RC for forecasting spatiotemporal dynamics.\cite{IngoPrivate}

Accounting for the translational symmetry somewhat improves the prediction horizon for the same training data size, \textit{i.e.}, the same value of $M$, as seen in Fig.~\ref{fig:Training1NG-RC}, which
displays a prediction horizon of 1.04 for the same initial condition used in Figs. \ref{fig:SingleNGRC} and \ref{fig:TrainingALLNG-RCs}. The mean prediction horizon for 100 different initial conditions increases by $\sim$29\% to 0.85  $\pm$ 0.17 MTU.

\subsection{\label{sec:PredictionAccuracy}Prediction accuracy comparison}

\begin{figure}[h]
\centering
\includegraphics[width=\linewidth]{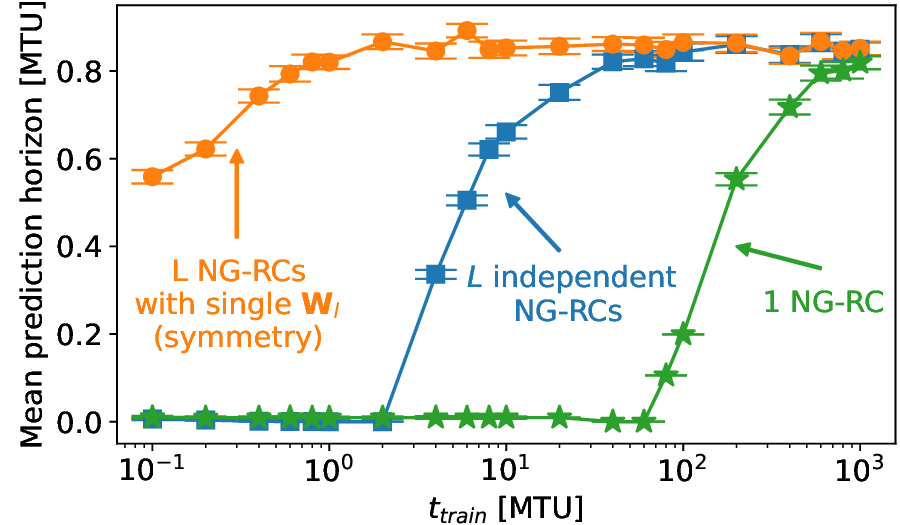}
\caption{Scaling of the mean prediction horizon with training time for the non-parallel model with a single NG-RC (green stars) and for the parallel NG-RCs using $L$ independent $\mathbf{\mathrm{W}}_l$'s (blue square) and using a single $\mathbf{\mathrm{W}}_l$ that respects translational symmetry (orange circles). Symbols represent the mean prediction horizon for 10 different training sets. For each training set we make predictions for 10 different initial conditions, total 100 predictions per point in the plot. Error bars represent the standard deviation of the mean over the 100 predictions. Parameters: $k=3$, $N_{nn}=2$ (for the parallel approaches) and $\alpha$ is optimized for each $t_{train}$ (see appendix \ref{app:optimization}).} 
\label{fig:MeanPred}
\end{figure}

As seen in Fig. \ref{fig:MeanPred}, the single NG-RC approach shows a mean prediction horizon near zero (poor prediction ability) for $t_{train}$ up to $\sim 60$ MTU where it starts to improve and reaches the value of 0.82 $\pm$ 0.15 MTU for $t_{train}=1,000$ (the maximum $t_{train}$ we explore in this work). For the  $L$ independent NG-RCs, the prediction horizon is near zero, begins to improve for $t_{train} \gtrsim 2$, and saturates above $t_{train} \gtrsim 40$, obtaining a similar value the single NG-RC approach with a training data set 25 times smaller. This makes the $L$ independent parallel NG-RCs to have a computational complexity $\approx 1.4\times10^3$ smaller than the single NG-RC. On the other hand, we obtain reasonable performance when respecting the translational symmetry for the smallest training time shown in the plot, and we obtain nearly the same prediction performance as the other approaches for $t_{train} \gtrsim 1$.  Thus, we see that we can reduce the observation time and computational cost by $\sim 1/L$, which is expected because the effective training time is $L$ times longer.   

\subsection{\label{sec:LowerDimensionCases}Lower dimensional cases for the Lorenz 96 model}

Here, we apply our model to lower dimensional cases of the extended Lorenz96 model and compare our results to previous research that use these simplified models.

\subsubsection{\label{sec:L8}Parallel NG-RC  model for $L=J=I=8$}

\begin{figure}[t]
\centering
\subfloat{%
	\includegraphics[width=\linewidth]{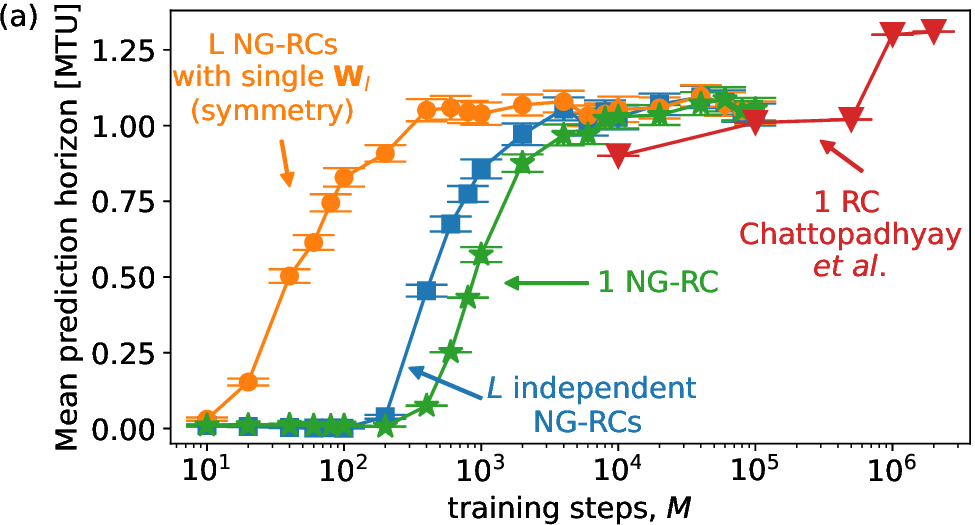}
	\label{fig:MeanPG_vs_train-L8_Parallel_NGRCs_a}
}\\
\subfloat{%
	\includegraphics[width=\linewidth]{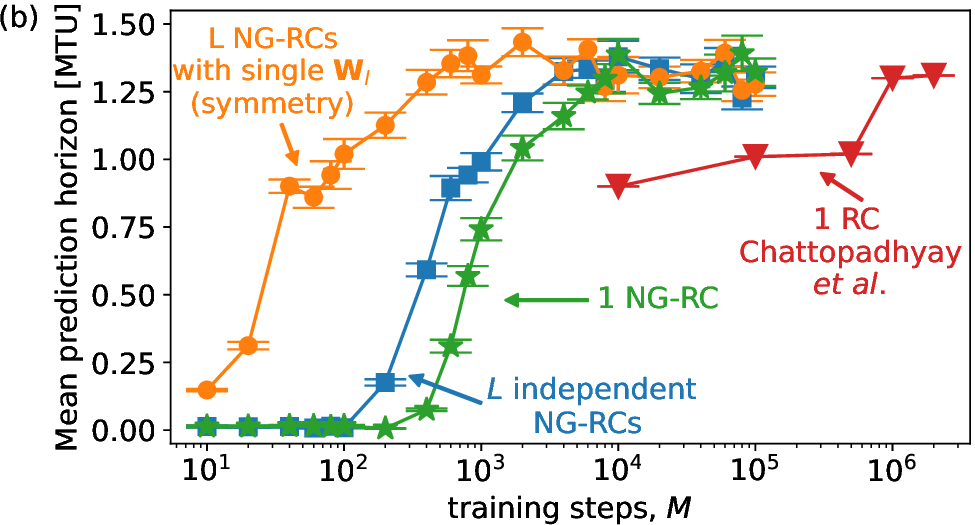}
	\label{fig:MeanPG_vs_train-L8_Parallel_NGRCs_b}
}
\caption{Mean prediction horizon for Lorenz96 system with $L=J=I=8$ as function of training steps $M$ for the non-parallel model with a single NG-RC (green stars) and for the parallel NG-RCs using $L$ independent $\mathbf{\mathrm{W}}_l$'s (blue square) and using a single $\mathbf{\mathrm{W}}_l$ that respects translational symmetry (orange circles). (a) Circle, square and star symbols represent the mean prediction horizon for 10 different training sets. For each training set we make predictions for 10 different initial conditions, totaling 100 predictions per point in the plot. The error bars represent the standard deviation of the mean over the 100 predictions. (b) Circle, square and star symbols represent the mean prediction horizon for the   training set (out of that 10 used in (a)) that returns the best prediction horizon. In both, the red down triangles represent Chattopadhyay {\textit{et al.}}'s results using a single RC. Parameters: $k=3$, $N_{nn}=2$ and $\alpha$ is optimized for each $M$.}
\label{fig:MeanPG_vs_train-L8_Parallel_NGRCs}
\end{figure}

First, we apply our parallel NG-RC approach ($N_{in}=5,N_{out}=1$) to the extended Lorenz96 system with $L=J=I=8$ and compare our results to our baseline method (single NG-RC) and to previous works. Figure \ref{fig:MeanPG_vs_train-L8_Parallel_NGRCs_a} shows the mean prediction horizon as function of the number of training steps $M$ for these cases. Here, we use  training steps $M$ rather than training time $t_{train}$ as the horizontal axis for a direct comparison to other works.  For our previous results presented above, the conversion is $t_{train}=M\delta t$.

\begin{figure}[t]
\centering
\subfloat{%
	\includegraphics[width=\linewidth]{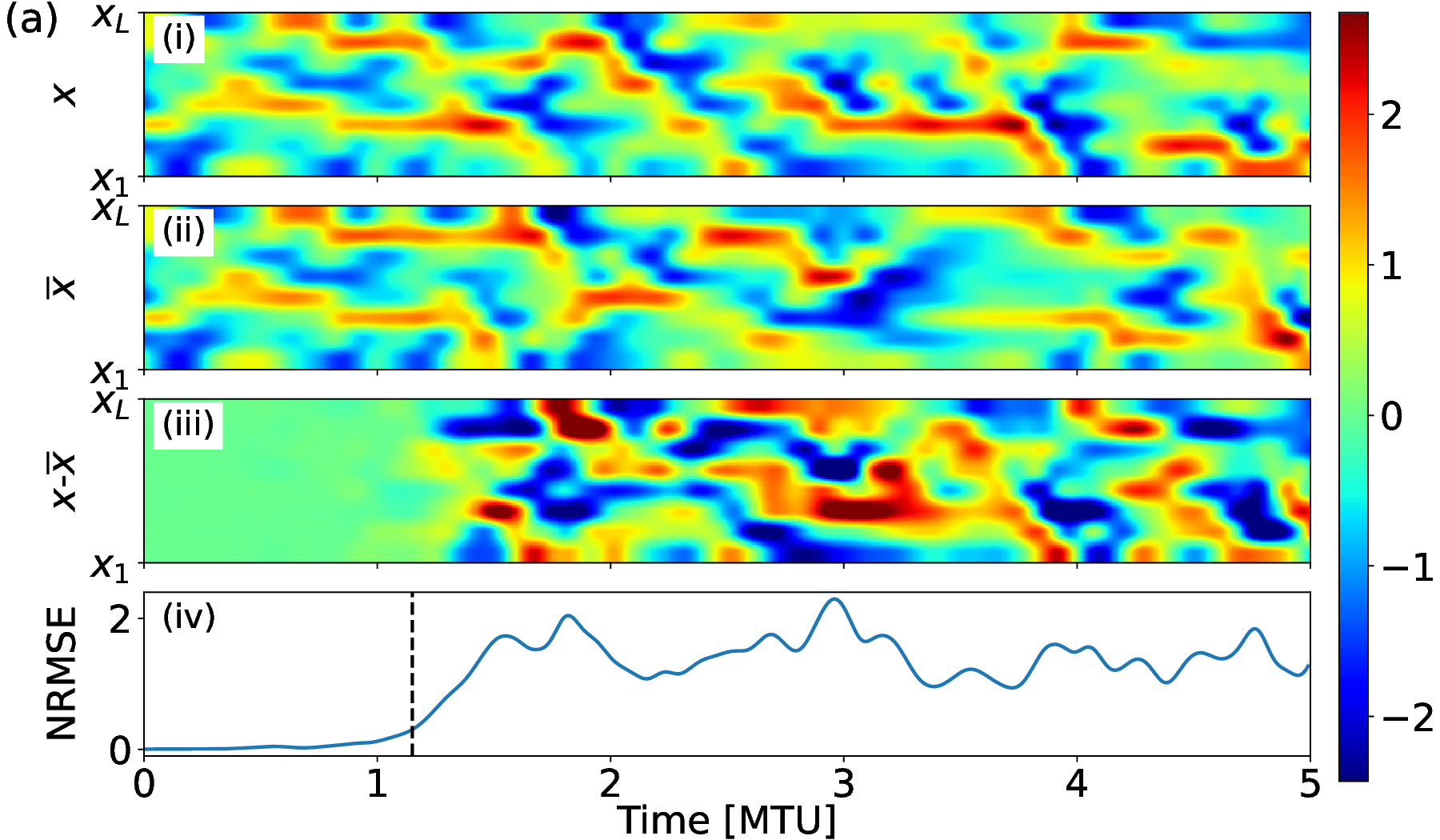} 
	\label{fig:fig:L8_Parallel_NGRCs_a}
}\\
\subfloat{%
	\includegraphics[width=\linewidth]{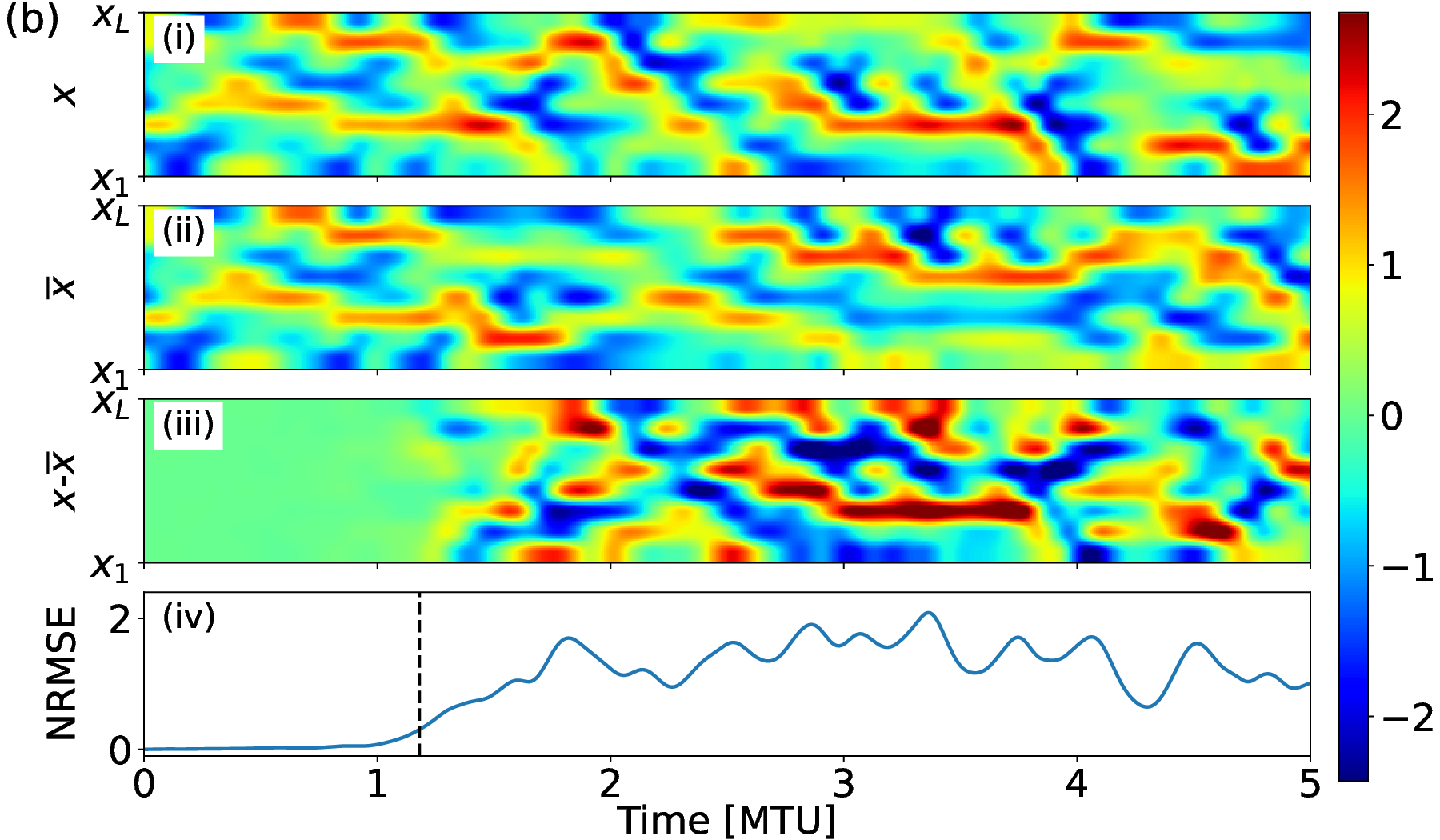}
	\label{fig:fig:L8_Parallel_NGRCs_b}
}\\
\subfloat{%
	\includegraphics[width=\linewidth]{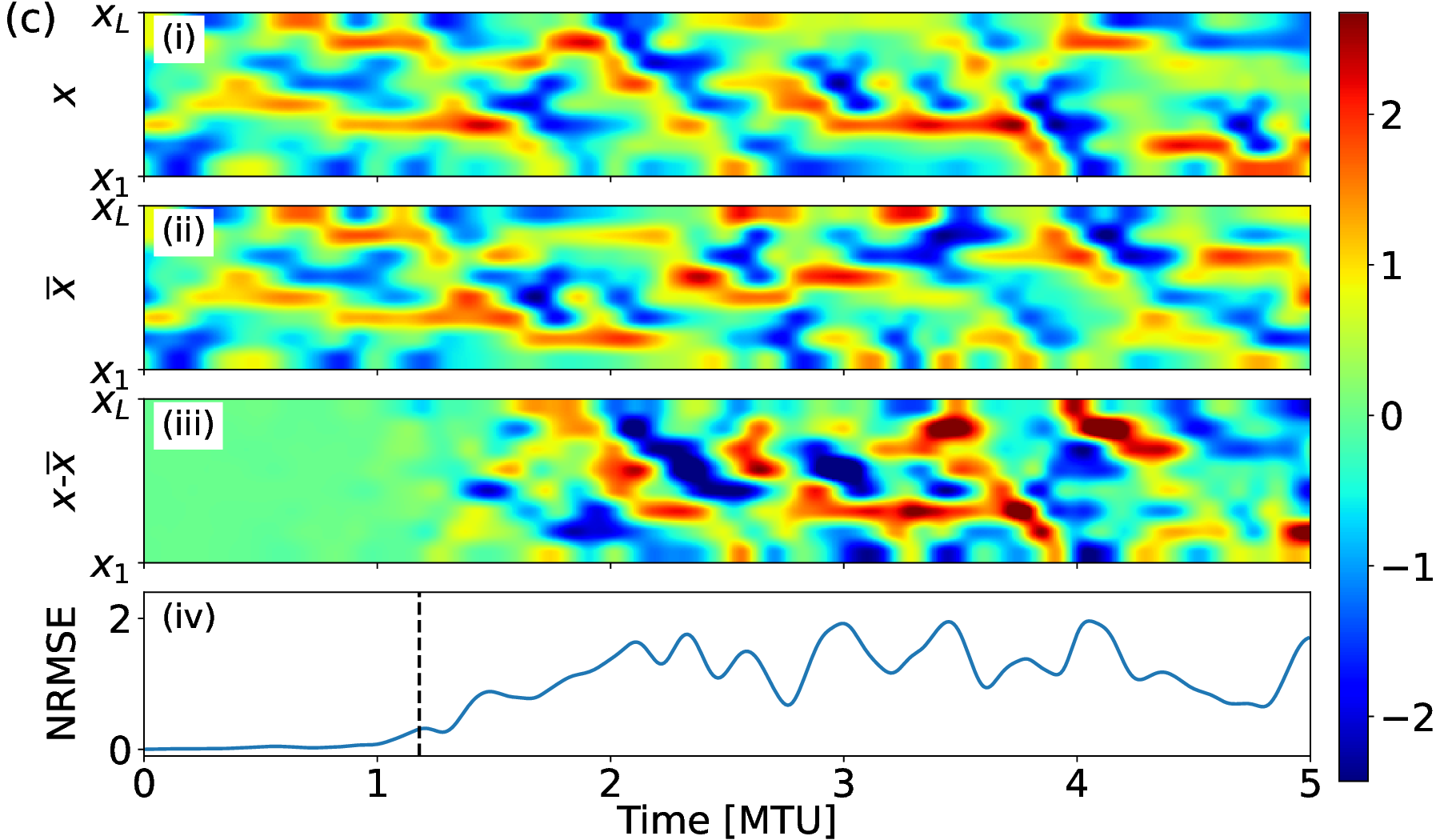}
	\label{fig:fig:L8_Parallel_NGRCs_c}
}
\caption{Typical prediction for the extended Lorenz96 system with $L=J=I=8$ using (a) a non-parallel scheme with a single NG-RC with $t_{train}=100$ MTU and $\alpha = 10^{-1}$, and using our parallel NG-RC approach with $N_{in}=5$ and $N_{out}=1$ for (b) $L$ independent $\mathbf{W}_l$'s with $t_{train}=40$ MTU and $\alpha = 10^{-3}$ and (c) a single $\mathbf{W}$ that respects translational symmetry with $t_{train}=4$ MTU and $\alpha = 4 \times 10^{-2}$. For the three panels: (i) Actual and (ii) predicted dynamics, (iii) difference between actual and predicted dynamics, and (iv) NRMSE. The vertical dashed line indicates the prediction horizon and $k=3$.}
\label{fig:L8_Parallel_NGRCs}
\end{figure}

When using a single NG-RC (green stars), we obtain a mean prediction horizon of  $1.03 \pm 0.39$ MTU for  $M=\gtrsim 10^4$ ($t_{train} \gtrsim 100$ MTU) where it starts to saturate. Here, the feature vector has $d_{total}= 325$ components. On the other hand, when using $L$ independent NG-RCs (blue squares), our model obtains a mean prediction horizon of $1.05 \pm 0.38$ MTU with $M=4,000$ corresponding to a training time of $ t_{train} = 40$ MTU. Here, each one of the $L$ parallel NG-RCs is trained individually with $M$ training points from the respective region using $d_{total}=136$ features, which results in a computational cost $\sim 2$ times smaller than the single NG-RC. Unsurprisingly, the  single NG-RC performs very similar to the $L$ parallel NG-RCs, as the number of spatial variables is low ($L=8$) and the single NG-RC input dimension $N_{in}=L$ contains fewer unrelated variables than the higher dimensional case shown in Fig. \ref{fig:MeanPred}. 
Finally, when using a single $\mathbf{W}$ that respects the translation symmetry (orange circles), we obtain a similar result (with no statistical difference) for $M=400$ ($t_{train}=4$ MTU), reducing the training time by a factor of $\sim$10 ($\sim L$) and a similar reduction in the computational cost. This reduction is expected because data from all $L$ spatial locations are concatenated to form a single training data set in this method with an effective size is $L$ times longer than $t_{train}=4$ MTU. Figure \ref{fig:L8_Parallel_NGRCs} shows typical predictions for the three cases discussed above.


%

Chattopadhyay {\textit{et al.}} \cite{Chattopadhyay2019} also predict the dynamics of this simplified extended Lorenz96 system using $N_{in}=N_{out}=L=8$ using a traditional RC with $d_{total}=5,000$ nodes (equal to the size of their feature vector) with $M=1\times10^4 - 2\times10^6$ training steps. Their results are shown in Fig. \ref{fig:MeanPG_vs_train-L8_Parallel_NGRCs_a} (red triangles). 
Their approach shows an improvement in performance when increasing $M$ to $5\times10^5$, where the mean prediction horizon is similar to our NG-RC approaches. We highlight that our parallel NG-RCs that respect translation symmetry (orange circles) presents similar results with a computational cost $\sim 2.1\times10^5$ smaller than their RC trained with  $M=5\times10^5$ data points (see Appendix \ref{app:ComputationalComplexity} for more details in computational complexity comparisons).     

Intriguingly, Chattopadhyay {\textit{et al.}}'s results show a step-like improvement when increasing the training steps to $M \geq 10^6$. They do not know the reason for this step and leave it for future investigations. It is important to notice that they use a single training data set in their work, whereas we average our results over multiple training data sets. One possible hypothesis for their observation is that the step improvement is related to the specific details of their training set.

To explore this possibility, we show the mean prediction horizon of our approaches using the best-performing single training data set in Fig. \ref{fig:MeanPG_vs_train-L8_Parallel_NGRCs_b}. We see that our best-performing training data set shows similar mean prediction horizon using less training data than the RC trained with $M \geq 10^6$ data points. Our results could be coincidental but points out one issue that may be responsible for explaining their observation.  The one take-away message is that it is important to average over training data sets to remove possible spurious effects on the prediction error due to the specific details of the training data set, which becomes more important for the short set sizes used here.

As another comparison of our work to past reports, Pyle {\it{et al.}} \cite{Pyle2021} also predict the same low-dimensional Lorenz96 system using a non-parallel ML scheme, except that they use an approach similar to the NG-RC with $k=1$ and all monomials up to quartic order. They use a single training data set of $M=5\times10^5$ training points, the same as Chattopadhyay \textit{et al.}, and obtain a mean prediction horizon of $1.60\pm0.53$ MTU. In terms of computational cost, our parallel approaches that does not (does) exploit the symmetry of the model is $2.1\times10^2$ ($2.1\times10^3$) less expensive than their approach.

\subsubsection{\label{sec:NoFineScale} Parallel NG-RC model for the Lorenz96 model without fine-scale variables with $L =40$, $J = I = 0$}


\begin{figure}[t]
\centering
\subfloat{%
	\includegraphics[width=\linewidth]{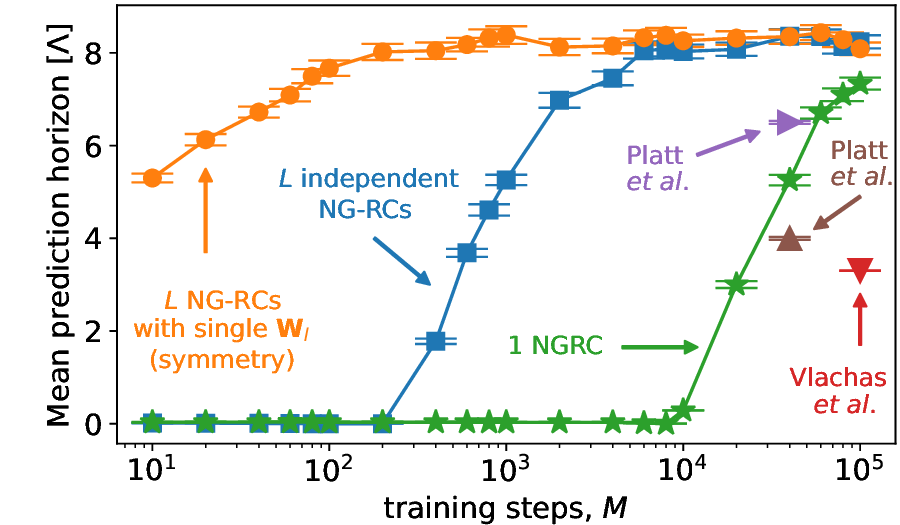}
}
\caption{Mean prediction horizon for Lorenz96 system with ($L=40$, $J=I=0$) as function of training steps $M$ for the non-parallel model with a single NG-RC (green stars) and for the parallel NG-RCs using $L$ independent $\mathbf{\mathrm{W}}_l$'s (blue square) and using a single $\mathbf{\mathrm{W}}_l$ that respects translational symmetry (orange circles). For these plots, the symbols represent the mean prediction horizon for 10 different training sets. For each training set we make predictions for 10 different initial conditions, totalizing 100 predictions per point in the plot. Error bars represent the standard deviation of the mean over the 100 predictions. The purple right triangle and brown up triangle represent Platt {\textit{et al.}}'s results for the same task using parallel RCs and single RC, respectively. The red down triangle represent the Vlachas  {\textit{et al.}}'s results for the same task using parallel RCs. Parameters: $k=3$, $N_{nn}=2$ and $\alpha$ is optimized for each $t_{train}$.}
\label{fig:L40_L40_IJ_0_vs_train}
\end{figure} 
  

Here, we use our approach to predict the dynamics of a simpler Lorenz96 model that does not include the fine spatiotemporal variables $y_{j,l}$ and $z_{i,j,l}$ ($J = I = 0$ - see Eq. \ref{eq:L96}) with $L=40$. 

First, we use our single NG-RC baseline model, a single NG-RC with $d_{total}= 7,381$ features. Similar to the predictions for the other Lorenz96 models shown in the previous sections, the single NG-RC only starts to improve its performance for a higher $M$ in comparison to the parallel approaches, as shown in Fig. \ref{fig:L40_L40_IJ_0_vs_train}, which displays the mean prediction horizon as function of the training steps $M$. The maximum mean prediction horizon for the single NG-RC is $7.3 \pm 1.3$ $\Lambda$ for $M=10^5$ ($t_{train}=1,000$ MTU), where $\Lambda = 1 / \lambda = 1/1.68$ MTU is the Lyapunov time with $L=40$ and $F=8$ (parameters used here).\cite{vlachas2019} 

When using $L$ independently trained NG-RCs (blue squares), the performance begins to improve for $M\gtrsim 200$ ($t_{train} \gtrsim 2$) and saturates for $M=6,000$ ($t_{train} = 60$), where the mean prediction horizon is $8.0 \pm 1.7$ $\Lambda$. Each parallel NG-RC has $d_{total}=136$ features, which makes the computational complex of this approach $1.2\times10^3$ times smaller than the single NG-RC method.  

\begin{figure}[t]
\centering
\subfloat{%
	\includegraphics[width=\linewidth]{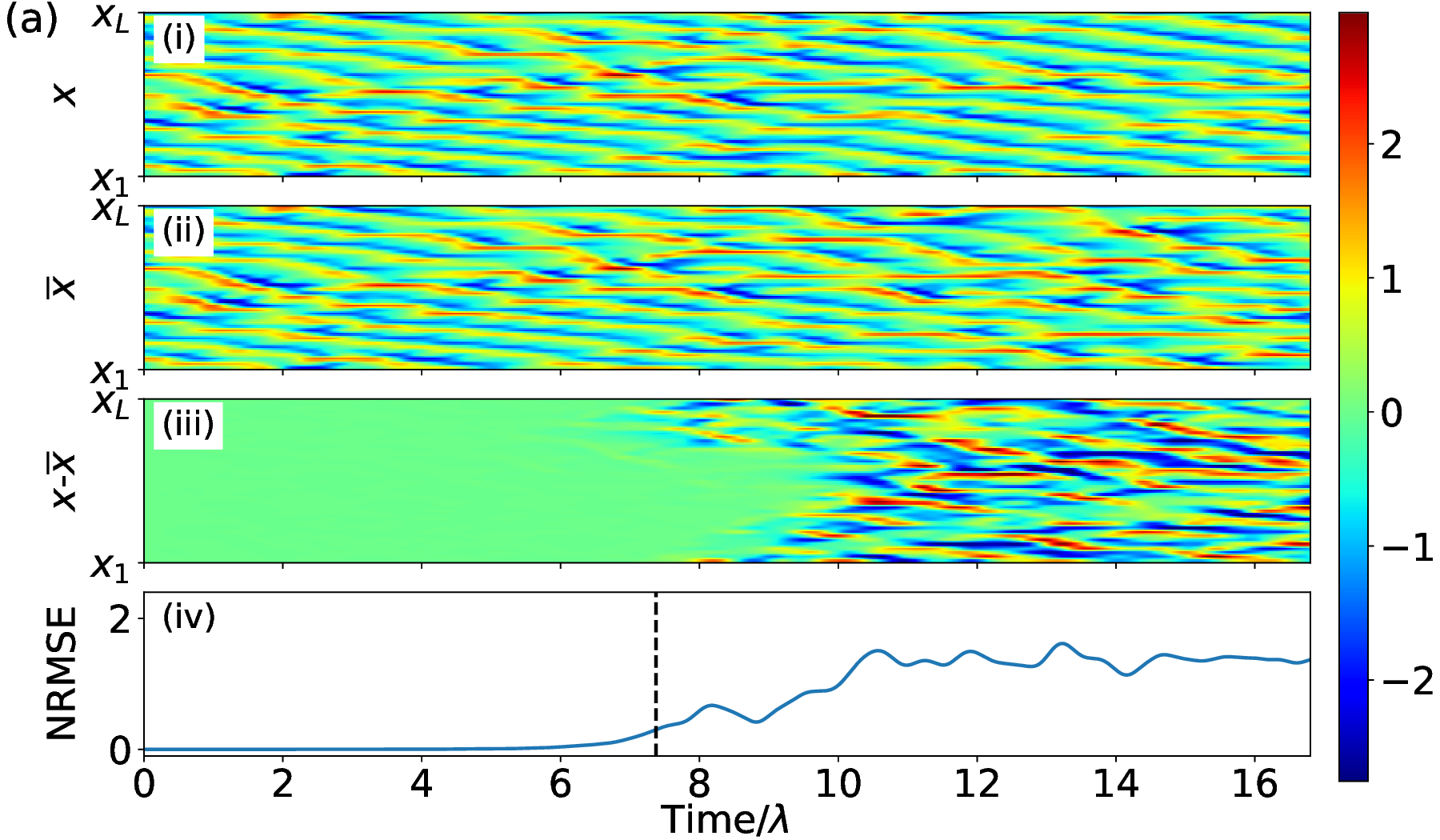} 
	\label{fig:L40_L40_IJ__a}
}\\
\subfloat{%
	\includegraphics[width=\linewidth]{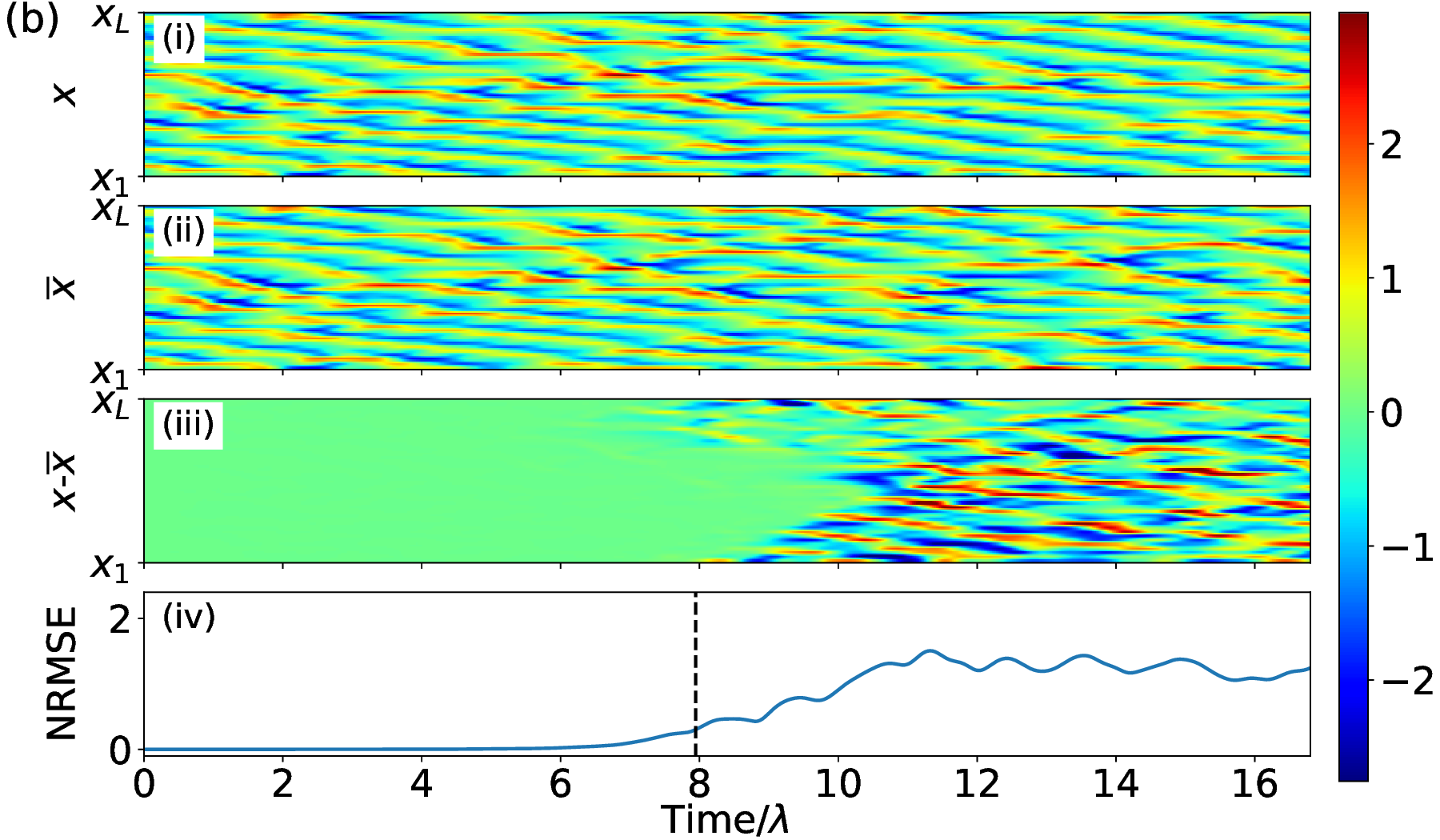}
	\label{fig:L40_L40_IJ__b}
}\\
\subfloat{%
	\includegraphics[width=\linewidth]{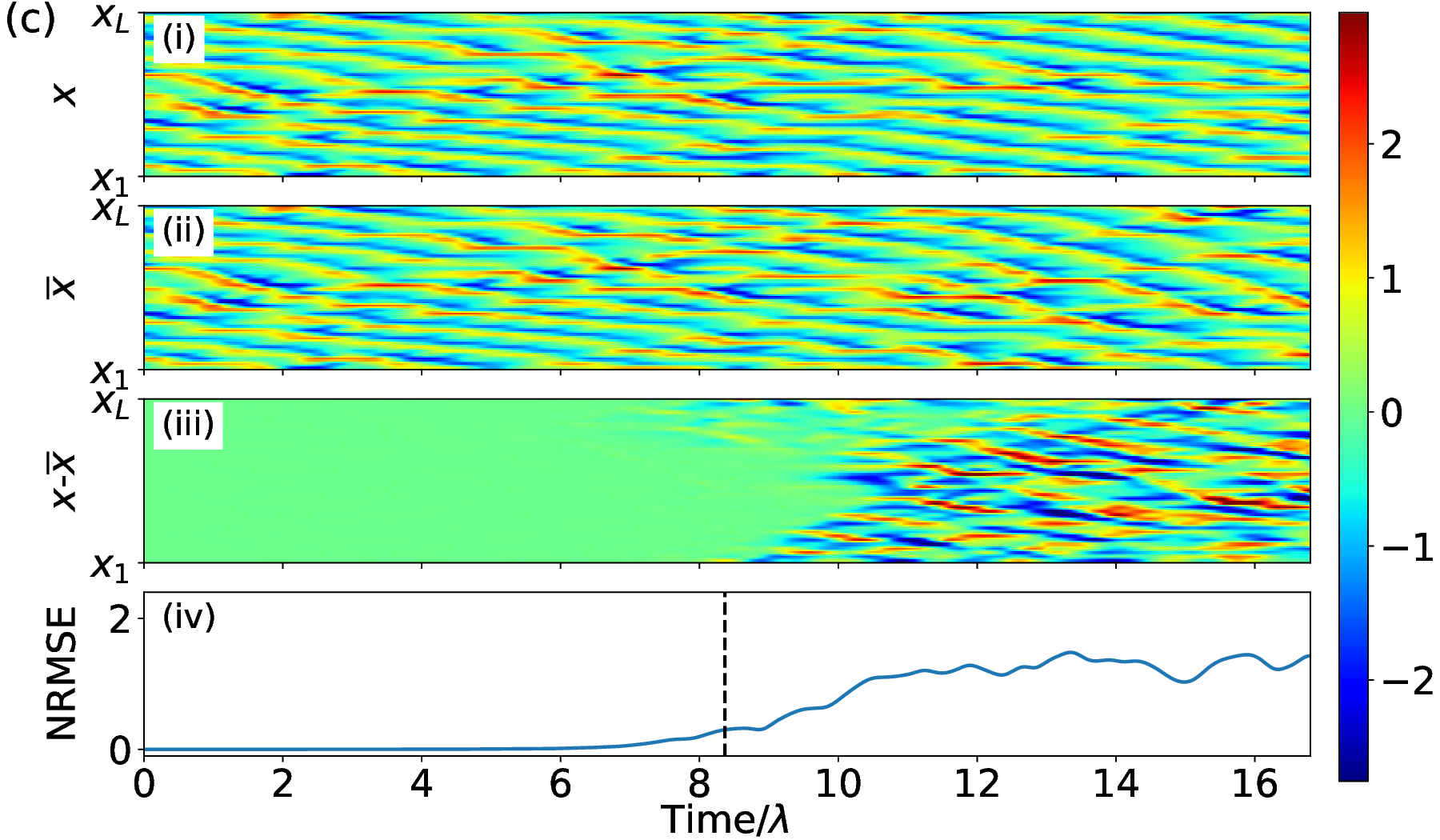}
	\label{fig:L40_L40_IJ__c}
}
\caption{Typical prediction for the extended Lorenz96 system with $L = 40$, and $J = I = 0$ using (a) a non-parallel scheme with a single NG-RC with $t_{train}=1000$ MTU,  and using our parallel NG-RC approach with $N_{in}=5$ and $N_{out}=1$ for (b) $L$ independent $\mathbf{W}_l$'s with $t_{train}=60$ MTU and (c) a single $\mathbf{W}$ that respects translational symmetry with $t_{train}=1$ MTU. For the three panels: (i) Actual and (ii) predicted dynamics, (iii) difference between actual and predicted dynamics, and (iv) NRMSE. The vertical dashed line indicates the prediction horizon. Parameters: $k = 3$ and  $\alpha = 1 \times 10^{-5}$.} 
\label{fig:L40_L40_IJ_0}
\end{figure}

In comparison, using a single $\mathbf{W}$ that respects the translation symmetry (orange circles), results in a mean prediction horizon of $7.7 \pm 1.7$ $\Lambda$ with $M=100$ (a training time of $t_{train}=1$ MTU). That is, we observe a similar prediction horizon using a factor of 60 smaller training data set in comparison to the $L$ independent NG-RCs.
Figure~\ref{fig:L40_L40_IJ_0} shows typical predictions the three cases. 

Vlachas {\textit{et al.}} \cite{vlachas2019} use a parallel RC scheme \cite{Pathak2018} to predict the same Lorenz96 model. They use 20 parallel RCs, each with $d_{total}=3,000$ nodes, $N_{in}=10>N_{out}=2$, and $M=100,000$. Their method obtains a mean prediction horizon of approximately $3.3$ $\Lambda$, represented by the red down triangle in Fig.  \ref{fig:L40_L40_IJ_0_vs_train}. When comparing to our approaches, we find that our $L$ independent NG-RCs model (blue squares) obtains a prediction horizon $2.4$ times longer using a training data set $17$ times smaller and with a computational complexity $\sim 4 \times 10^3$ times shorter than their result. When considering the translational symmetry (orange circles), our approach obtains the same $2.4$ improvement factor on the prediction horizon, but with a training data set $10^3$ times smaller and with a computational complexity $\sim 2.4 \times 10^5$ times shorter than their approach.  

Recently, Platt {\it{et al.}} \cite{Platt2022} optimized the parallel RC architecture to obtain a mean prediction horizon twice as long as Vlachas {\textit{et al.}} using $N_{in}=6>N_{out}=2$ with smaller RCs (each with $d_{total}=720$ nodes) and training data set ($M=40,000$). Their result  is represented by the purple right triangle on Fig.~\ref{fig:L40_L40_IJ_0_vs_train}. Our parallel model with independent NG-RCs obtains slightly better performance in the mean prediction horizon with a computational complexity $\sim 10^2$  shorter using $\sim 7\times$ less training data than Platt {\textit{et al.}}. Considering the translational symmetry, our model presents a computational complexity $5.6 \times 10^3$ smaller using $4\times 10^2$ less training data. See Appendix \ref{app:ComputationalComplexity} for more details on the computational complexity comparison. Lastly, Platt {\textit{et al.}} also use a single RC with $d_{total}=6,000$ nodes to predict the same system. While this model presents a worse performance in comparison to the three NG-RC approaches presented here, it surprisingly outperform the parallel RC model by Vlachas {\textit{et al.}}, indicating an important improvement due to the optimizations done by Platt {\textit{et al.}}.


\section{\label{sec:Discussion_Conclusion}Discussions and Conclusions}

We emphasize that parallel machine learning architectures provide high-efficiency prediction of high-dimensional spatiotemporal dynamical system. Partitioning the learning system is small subsystems, each of which can be predicted using a smaller ML model unit, provides better predictions and is less computational expensive than using a single model to predict the entire system. 

In our proposed method, we maximize the parallelism and let each parallel ML model unit predict a single variable of the system. Thus, there are as many parallel units as the number of predicted variables in the system (here the slow variables of the Lorenz96 model). Furthermore, this parallel approach lends itself to parallelization methods using graphical processor units or multi-processor computer clusters. 
 
We show that our parallel scheme composed by independently trained NG-RCs outperforms a non-parallel model composed by a single larger NG-RC. For the higher dimensional Lorenz96 system addressed in this work ($L=36,I=J=10$), the parallelization provides a $\sim 1.4\times10^3$ improvement factor in the computational cost while obtaining similar prediction accuracy using 25 times smaller training data set than the non-parallel architecture. We further decrease both training data and computational cost by another factor of $\sim L$ by addressing the system translational symmetry, which is sometimes present in spatiotemporal systems with cyclic boundary conditions. 

We also predict the dynamics of lower-dimensional Lorenz96-like models to compare our results to previous research. We demonstrate that our method is more accurate, or can be training with less data and computational cost, or both. For the extended Lorenz96 with $L=I=J=8$, we show that our parallel approach obtains similar results to a traditional RC implemented by Chattopadhyay {\textit{et al.}} \cite{Chattopadhyay2019}, but with a computational cost up to $\sim 2.1\times10^4$ smaller using up to $1.2\times10^2$ less training data. When considering the translational symmetry, these numbers improve by another factor of 10. Chattopadhyay {\textit{et al.}} also implement a deep learning network for this problem, but the deep learning method demonstrates no accuracy improvement and requires more computation time. 

We also use our approach to predict the Lorenz96 system for $L=40,I=J=0$, where we obtain better results with a training data set up to $10^3$ times smaller and computational costs up to $2.4\times10^5$ smaller than parallel implementations of traditional RCs implemented by Vlachas {\textit{et al.}} \cite{vlachas2019}. We also obtain better results  than parallel implementations of traditional RCs implemented  Platt {\it{et al.}} \cite{Platt2022} using up to $4\times10^2$ times less training data and a computational costs up to $5.6\times10^3$ smaller.

The low computational cost, less training data requirement and fewer optimizable parameters of the NG-RC in comparison to other ML approaches allow us to implement our parallel architecture in a standard desktop computer and obtain sub-seconds training and prediction times. 
To give an absolute scale for the computational cost for our approach, we produce the results for this paper using Python 3.7.6, NumPy 1.19.15 and scikit-learn 0.24.2 on an x86-64 CPU running Windows 10. For the results presented in Figs. \ref{fig:TrainingALLNG-RCs} and \ref{fig:Training1NG-RC}, the computation time for training all $L=36$ NG-RCs with $M=1,000$ data points is $55 \pm 1$ ms while the runtime for predicting one time step is $394 \pm 3$ $\mu$s, or $10.9 \pm 0.1$ $\mu$s per spatial location per step.

Future directions of our research include hybrid approaches using NG-RCs that combine model-generated and experimentally observed data such as those explored using a traditional RC. \cite{Wikner2020,Fan2020,Wikner2021,Arcomano2022} Also, our method should generalize to two- and three-dimensional fluid dynamics problems where an even greater reduction in the required data set size is anticipated.  Combined with our approach of one-step-ahead prediction, a coarser spatiotemporal grid can be used, offering the possibility of greatly speeding up spatiotemporal simulations. 

\begin{acknowledgments}
We gratefully acknowledge discussions of this work with E. Bollt and the financial support of the Air Force Office of Scientific Research, Contract \#FA9550-20-1-0177.
\end{acknowledgments}

\section*{Data Availability Statement}

The data that support the findings of this study are available from the corresponding author upon reasonable request.

\appendix

\section{\label{app:optimization}Ridge regression Parameter Optimization}

We train our model using a supervised learning algorithm where the input data ${\bf{x}}$ and the desired output ${\bf{y}}$  are known in advance for the entire training period.  Here, ${\bf{y}}$ is evaluated at time $t_{m+1}$, whereas ${\bf{x}}$ is evaluated at time $t_m$ because we are making a next-step-ahead prediction. We use Ridge regression to find the matrix of weights ${\mathbf{W}}$ that minimizes  
 \begin{equation}
||\overline {\bf{y}}-{\bf{y}}||^2 + \alpha||{\mathbf{W}}||^2, 
\label{eq:Ridge}
 \end{equation}
\noindent where $\overline {\bf{y}}={\mathbf{W}}{\mathcal O}_{total}$ is the model output, ${\mathcal O}_{total}$ is the feature vector obtained from the input data  ${\bf{x}}$ and $\alpha$ is the Ridge parameter. Here, $||\cdot||$ represents the L2-norm. When $\alpha$ is zero, Ridge regression reduces to regular least-square regression. The Ridge parameter adds a penalty term to prevent overfitting.   

\begin{figure*}[t]
\centering

\subfloat{%
	\includegraphics[width=\linewidth]{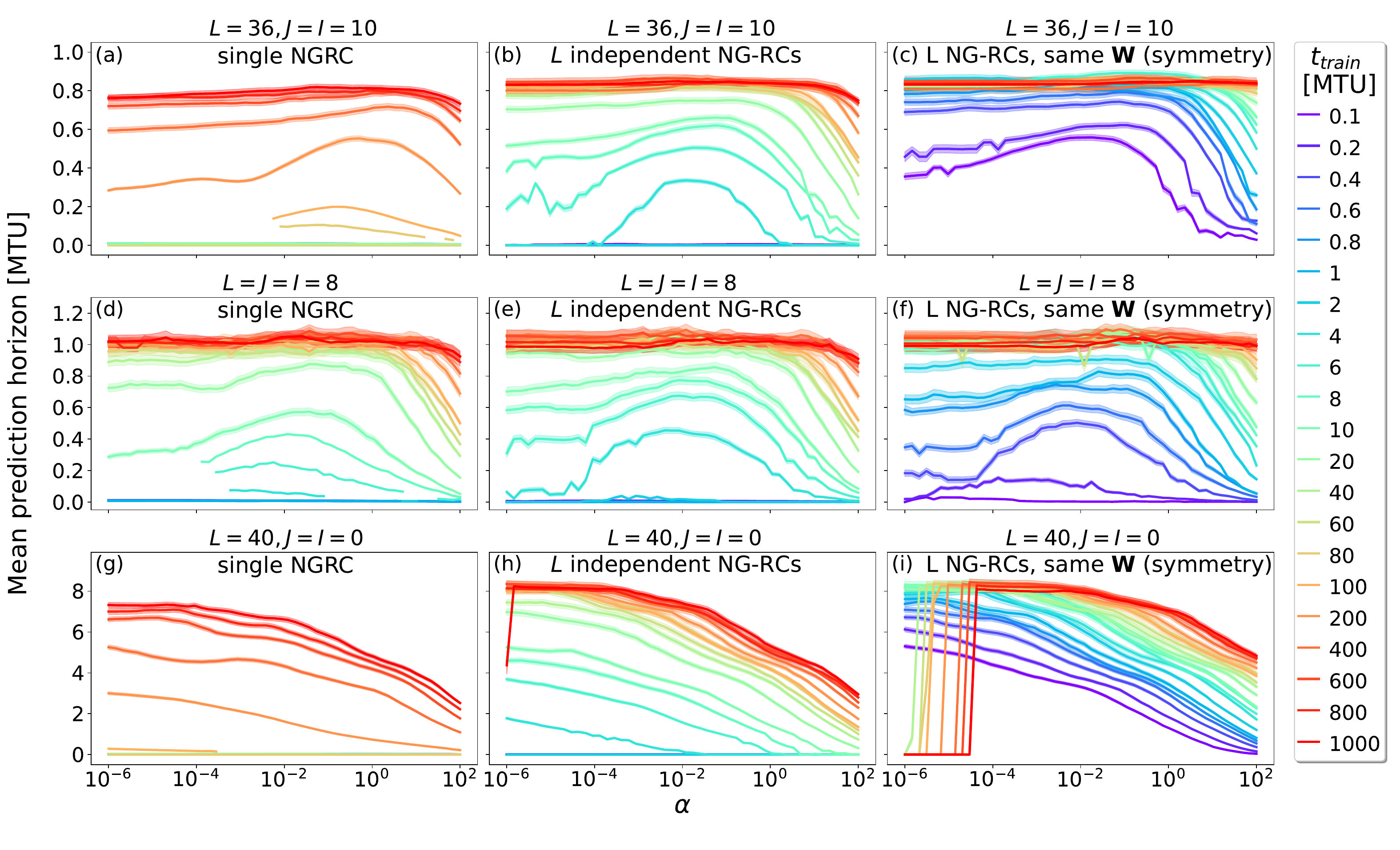}%
	\label{fig:MeanPH_vs_alpha_diff_training_times_a}%
}
\begin{minipage}[t]{0\textwidth}
	\subfloat{%
	\label{fig:MeanPH_vs_alpha_diff_training_times_b}%
	}
\end{minipage}
\begin{minipage}[t]{0\textwidth}
	\subfloat{%
	\label{fig:MeanPH_vs_alpha_diff_training_times_c}%
	}
\end{minipage}
\begin{minipage}[t]{0\textwidth}
	\subfloat{%
	\label{fig:MeanPH_vs_alpha_diff_training_times_d}%
	}
\end{minipage}
\begin{minipage}[t]{0\textwidth}
	\subfloat{%
	\label{fig:MeanPH_vs_alpha_diff_training_times_e}%
	}
\end{minipage}
\begin{minipage}[t]{0\textwidth}
	\subfloat{%
	\label{fig:MeanPH_vs_alpha_diff_training_times_f}%
	}
\end{minipage}
\begin{minipage}[t]{0\textwidth}
	\subfloat{%
	\label{fig:MeanPH_vs_alpha_diff_training_times_g}%
	}
\end{minipage}
\begin{minipage}[t]{0\textwidth}
	\subfloat{%
	\label{fig:MeanPH_vs_alpha_diff_training_times_h}%
	}
\end{minipage}
\begin{minipage}[t]{0\textwidth}
	\subfloat{%
	\label{fig:MeanPH_vs_alpha_diff_training_times_i}%
	}
\end{minipage}
\caption{Ridge parameter optimization for the results shown in Figs. \ref{fig:MeanPred}, \ref{fig:MeanPG_vs_train-L8_Parallel_NGRCs} and \ref{fig:L40_L40_IJ_0_vs_train}. Mean prediction horizon as function of the Ridge parameter $\alpha$ for different training times (see legend color code) for the Lorenz96 system with $L=36,J=I=10$ ((a)-(c)), $L=J=I=8$ ((d)-(f)) and  $L=40,J=I=0$ ((g)-(i)). For each case, optimizations for a single NG-RC, $L$ independent NG-RCs and $L$ NG-RCs using translational symmetry are presented. The colored area around the curves represent the standard deviation of the mean.}    
\label{fig:MeanPH_vs_alpha_diff_training_times}
\end{figure*}

We use a grid-search procedure to find the optimal $\alpha$ that maximizes the mean prediction horizon for each case shown in previous sections.  Figure \ref{fig:MeanPH_vs_alpha_diff_training_times} shows the optimization results for all cases shown in Figs. \ref{fig:MeanPred}, \ref{fig:MeanPG_vs_train-L8_Parallel_NGRCs} and \ref{fig:L40_L40_IJ_0_vs_train}. For each $\alpha$, we calculate the mean prediction horizon for 10 different training sets.   For each training set we perform predictions for 10 different initial conditions, totaling 100 predictions per $\alpha$.  We repeat this process for different training times $t_{train}$. 
It is seen that for $L=36,J=I=10$ (Figs.\ref{fig:MeanPH_vs_alpha_diff_training_times_a}-\ref{fig:MeanPH_vs_alpha_diff_training_times_c}) and for $L=J=I=8$ (Figs.\ref{fig:MeanPH_vs_alpha_diff_training_times_d}-\ref{fig:MeanPH_vs_alpha_diff_training_times_f}), the three NG-RC algorithms are not sensitive to the value of $\alpha$ for higher training times. On the other hand, for $L=40,J=I=0$ (Figs.\ref{fig:MeanPH_vs_alpha_diff_training_times_g}-\ref{fig:MeanPH_vs_alpha_diff_training_times_i}), the three algorithms show a trend to have a worse performance for higher values of $\alpha$.

\section{\label{app:ComputationalComplexity}Computational Complexity}

Here, we provide an estimation of the computational complexity our parallel NG-RC approach in comparison to the other RC-based approaches mentioned above. The main contribution to the computational complexity for both the NG-RC and the regular RC is performing the Ridge regression, which scales as $\mathcal{O}(Md_{total}^2)$ for $M$ training points and $d_{total}$ features. The comparison for the extended Lorenz96 system with ($L=36,J=I=10$), ($L=J=I=8$) and ($L=40,J=I=0$) are shown in Tables \ref{tab0}, \ref{tab1} and \ref{tab2}, respectively. Note that, for the case where we train a single $\mathbf{W}$ respecting the translational symmetry (first row of each table), the number of training points is multiplied by the number of spatial locations $L$ to reflect the training data concatenation as discussed in previous sections.    

In our analysis for the RC complexity, we do not account for the cost of multiplying the nodes states with the adjacency matrix that represent the network because these RC approaches use neural networks with sparse connectivity (we assume this fact when it is not stated). Also, we do not take into account special function evaluation costs, such as the hyperbolic tangent present in the traditional RC. For the NG-RC, we do not take into account the computational cost of the feature vector creation that happens before training.

\begin{table*}[t!]
  \centering
  \begin{tabular}{|c|c|c|c|c|c|c|c|}
  \hline
     & ML model & M & $d_{total}$  & $N_{in}$& $N_{out}$ &  Parallel units trained & Speed up  \\ \hline

    Our approach - Respecting symmetry  
    & NG-RC
    & 100 $\times$ 36  
    & 136
    & 5 
    & 1
    & -
    & - 
    \\ \hline
   
    Our approach - $L$ independent NG-RCs  
    & NG-RC
    & 4,000  
    & 136
    & 5 
    & 1
    & 36
    & 40 
    \\ \hline
 
     Our approach - Single NG-RC  
    & NG-RC
    & 100,000  
    & 5995
    & 36 
    & 36
    & -
    & $5.4\times10^4$ 
    \\ \hline

  \end{tabular}
  \caption{Training complexity comparison of different ML approaches for prediction of the extend Lorenz96 system with $L=36,J=I=10$.  
  }\label{tab0}
\end{table*}

\begin{table*}[t!]
  \centering
  \begin{tabular}{|c|c|c|c|c|c|c|c|}
  \hline
     & ML model & M & $d_{total}$  & $N_{in}$& $N_{out}$ &  Parallel units trained & Speed up  \\ \hline

    Our approach - Respecting symmetry  
    & NG-RC
    & 400 $\times$ 8  
    & 136
    & 5 
    & 1
    & -
    & - 
    \\ \hline
   
    Our approach - $L$ independent NG-RCs  
    & NG-RC
    & 4,000  
    & 136
    & 5 
    & 1
    & 8
    & 10 
    \\ \hline
 
     Our approach - Single NG-RC  
    & NG-RC
    & 10,000  
    & 325
    & 8 
    & 8
    & -
    & 18 
    \\ \hline
   
    Chattopadhyay {\it{et al.}} \cite{Chattopadhyay2019}  
    & RC
    & 500,000  
    & 5,000
    & 8 
    & 8
    & -
    & $2.1 \times 10^5$ 
    \\ \hline
 
    Pyle {\it{et al.}} \cite{Pyle2021}  
    & NG-RC
    & 500,000  
    & 495
    & 8 
    & 8
    & -
    & $2.1 \times 10^3$ 
    \\ \hline

  \end{tabular}
  \caption{Training complexity comparison of different ML approaches for prediction of the extend Lorenz96 system with $L=J=I=8$.  
  }\label{tab1}
\end{table*}

\begin{table*}[t!]
  \centering
  \begin{tabular}{|c|c|c|c|c|c|c|c|}
  \hline
     & ML model & M & $d_{total}$  & $N_{in}$& $N_{out}$ &  Parallel units trained & Speed up  \\ \hline

    Our approach - Respecting symmetry     
    & NG-RC
    & 100 $\times$ 40  
    & 136
    & 5 
    & 1
    & -
    & - 
    \\ \hline
   
    Our approach - $L$ independent NG-RCs    
    & NG-RC
    & 6,000  
    & 136
    & 5 
    & 1
    & 40
    & 60 
    \\ \hline
   
    Our approach - Single NG-RC  
    & NG-RC
    & 100,000  
    & 7381
    & 40 
    & 40
    & -
    & $7\times10^4$ 
    \\ \hline
   
    Vlachas {\it{et al.}} \cite{vlachas2019}  
    & RC
    & 100,000  
    & 3,000
    & 10 
    & 2
    & 20
    & $2.4 \times 10^5$ 
    \\ \hline
 
    Platt {\it{et al.}} \cite{Platt2022}  
    & RC
    & 40,000 
    & 720
    & 6 
    & 2
    & 20
    & $5.6 \times 10^3$ 
    \\ \hline

  \end{tabular}
  \caption{Training complexity comparison of different ML approaches for prediction of the extend Lorenz96 system with $L=40$ and $J=I=0$.  
  }\label{tab2}
\end{table*}
%


\begin{thebibliography}{32}%
\makeatletter
\providecommand \@ifxundefined [1]{%
 \@ifx{#1\undefined}
}%
\providecommand \@ifnum [1]{%
 \ifnum #1\expandafter \@firstoftwo
 \else \expandafter \@secondoftwo
 \fi
}%
\providecommand \@ifx [1]{%
 \ifx #1\expandafter \@firstoftwo
 \else \expandafter \@secondoftwo
 \fi
}%
\providecommand \natexlab [1]{#1}%
\providecommand \enquote  [1]{``#1''}%
\providecommand \bibnamefont  [1]{#1}%
\providecommand \bibfnamefont [1]{#1}%
\providecommand \citenamefont [1]{#1}%
\providecommand \href@noop [0]{\@secondoftwo}%
\providecommand \href [0]{\begingroup \@sanitize@url \@href}%
\providecommand \@href[1]{\@@startlink{#1}\@@href}%
\providecommand \@@href[1]{\endgroup#1\@@endlink}%
\providecommand \@sanitize@url [0]{\catcode `\\12\catcode `\$12\catcode
  `\&12\catcode `\#12\catcode `\^12\catcode `\_12\catcode `\%12\relax}%
\providecommand \@@startlink[1]{}%
\providecommand \@@endlink[0]{}%
\providecommand \url  [0]{\begingroup\@sanitize@url \@url }%
\providecommand \@url [1]{\endgroup\@href {#1}{\urlprefix }}%
\providecommand \urlprefix  [0]{URL }%
\providecommand \Eprint [0]{\href }%
\providecommand \doibase [0]{https://doi.org/}%
\providecommand \selectlanguage [0]{\@gobble}%
\providecommand \bibinfo  [0]{\@secondoftwo}%
\providecommand \bibfield  [0]{\@secondoftwo}%
\providecommand \translation [1]{[#1]}%
\providecommand \BibitemOpen [0]{}%
\providecommand \bibitemStop [0]{}%
\providecommand \bibitemNoStop [0]{.\EOS\space}%
\providecommand \EOS [0]{\spacefactor3000\relax}%
\providecommand \BibitemShut  [1]{\csname bibitem#1\endcsname}%
\let\auto@bib@innerbib\@empty
\bibitem [{\citenamefont {Winfree}(1987)}]{winfree1987time}%
  \BibitemOpen
  \bibfield  {author} {\bibinfo {author} {\bibfnamefont {A.}~\bibnamefont
  {Winfree}},\ }\href {https://books.google.com/books?id=zpuFQgAACAAJ} {\emph
  {\bibinfo {title} {When Time Breaks Down}}}\ (\bibinfo  {publisher}
  {Princeton University Press},\ \bibinfo {year} {1987})\BibitemShut {NoStop}%
\bibitem [{\citenamefont {Illing}, \citenamefont {Gauthier},\ and\
  \citenamefont {Roy}(2007)}]{ILLING2007bib}%
  \BibitemOpen
  \bibfield  {author} {\bibinfo {author} {\bibfnamefont {L.}~\bibnamefont
  {Illing}}, \bibinfo {author} {\bibfnamefont {D.~J.}\ \bibnamefont
  {Gauthier}},\ and\ \bibinfo {author} {\bibfnamefont {R.}~\bibnamefont
  {Roy}},\ }\href
  {https://doi.org/https://doi.org/10.1016/S1049-250X(06)54010-8} {\emph
  {\bibinfo {title} {Controlling Optical Chaos, Spatio-Temporal Dynamics, and
  Patterns}}},\ edited by\ \bibinfo {editor} {\bibfnamefont {P.}~\bibnamefont
  {Berman}}, \bibinfo {editor} {\bibfnamefont {C.}~\bibnamefont {Lin}},\ and\
  \bibinfo {editor} {\bibfnamefont {E.}~\bibnamefont {Arimondo}},\ \bibinfo
  {series} {Advances In Atomic, Molecular, and Optical Physics}, Vol.~\bibinfo
  {volume} {54}\ (\bibinfo  {publisher} {Academic Press},\ \bibinfo {year}
  {2007})\ pp.\ \bibinfo {pages} {615--697}\BibitemShut {NoStop}%
\bibitem [{\citenamefont {Holmes}, \citenamefont {Lumley},\ and\ \citenamefont
  {Berkooz}(1996)}]{holmes_lumley_berkooz_1996}%
  \BibitemOpen
  \bibfield  {author} {\bibinfo {author} {\bibfnamefont {P.}~\bibnamefont
  {Holmes}}, \bibinfo {author} {\bibfnamefont {J.~L.}\ \bibnamefont {Lumley}},\
  and\ \bibinfo {author} {\bibfnamefont {G.}~\bibnamefont {Berkooz}},\ }\href
  {https://doi.org/10.1017/CBO9780511622700} {\emph {\bibinfo {title}
  {Turbulence, Coherent Structures, Dynamical Systems and Symmetry}}},\
  Cambridge Monographs on Mechanics\ (\bibinfo  {publisher} {Cambridge
  University Press},\ \bibinfo {year} {1996})\BibitemShut {NoStop}%
\bibitem [{\citenamefont {Lorenz}(1996)}]{Lorenz1996}%
  \BibitemOpen
  \bibfield  {author} {\bibinfo {author} {\bibfnamefont {E.}~\bibnamefont
  {Lorenz}},\ }\bibfield  {title} {\enquote {\bibinfo {title} {Predictability:
  a problem partly solved},}\ }in\ \href@noop {} {\emph {\bibinfo {booktitle}
  {Proc. Seminar on Predictability}}},\ Vol.~\bibinfo {volume} {1}\ (\bibinfo
  {organization} {ECMWF},\ \bibinfo {address} {Reading, Berkshire, United
  Kingdom},\ \bibinfo {year} {1996})\ pp.\ \bibinfo {pages} {1--18}\BibitemShut
  {NoStop}%
\bibitem [{\citenamefont {Wilks}(2005)}]{Wilks2005}%
  \BibitemOpen
  \bibfield  {author} {\bibinfo {author} {\bibfnamefont {D.~S.}\ \bibnamefont
  {Wilks}},\ }\bibfield  {title} {\enquote {\bibinfo {title} {Effects of
  stochastic parametrizations in the lorenz '96 system},}\ }\href
  {https://doi.org/https://doi.org/10.1256/qj.04.03} {\bibfield  {journal}
  {\bibinfo  {journal} {Q. J. R. Meteorol. Soc.}\ }\textbf {\bibinfo {volume}
  {131}},\ \bibinfo {pages} {389--407} (\bibinfo {year} {2005})}\BibitemShut
  {NoStop}%
\bibitem [{\citenamefont {Chattopadhyay}, \citenamefont {Hassanzadeh},\ and\
  \citenamefont {Subramanian}(2020)}]{Chattopadhyay2019}%
  \BibitemOpen
  \bibfield  {author} {\bibinfo {author} {\bibfnamefont {A.}~\bibnamefont
  {Chattopadhyay}}, \bibinfo {author} {\bibfnamefont {P.}~\bibnamefont
  {Hassanzadeh}},\ and\ \bibinfo {author} {\bibfnamefont {D.}~\bibnamefont
  {Subramanian}},\ }\bibfield  {title} {\enquote {\bibinfo {title} {Data-driven
  predictions of a multiscale lorenz 96 chaotic system using machine-learning
  methods: reservoir computing, artificial neural network, and long short-term
  memory network},}\ }\href {https://doi.org/10.5194/npg-27-373-2020}
  {\bibfield  {journal} {\bibinfo  {journal} {Nonlinear Process. Geophys.}\
  }\textbf {\bibinfo {volume} {27}},\ \bibinfo {pages} {373--389} (\bibinfo
  {year} {2020})}\BibitemShut {NoStop}%
\bibitem [{\citenamefont {Pyle}\ \emph {et~al.}(2021)\citenamefont {Pyle},
  \citenamefont {Jovanovic}, \citenamefont {Subramanian}, \citenamefont
  {Palem},\ and\ \citenamefont {Patel}}]{Pyle2021}%
  \BibitemOpen
  \bibfield  {author} {\bibinfo {author} {\bibfnamefont {R.}~\bibnamefont
  {Pyle}}, \bibinfo {author} {\bibfnamefont {N.}~\bibnamefont {Jovanovic}},
  \bibinfo {author} {\bibfnamefont {D.}~\bibnamefont {Subramanian}}, \bibinfo
  {author} {\bibfnamefont {K.~V.}\ \bibnamefont {Palem}},\ and\ \bibinfo
  {author} {\bibfnamefont {A.~B.}\ \bibnamefont {Patel}},\ }\bibfield  {title}
  {\enquote {\bibinfo {title} {Domain-driven models yield better predictions at
  lower cost than reservoir computers in lorenz systems},}\ }\href
  {https://doi.org/10.1098/rsta.2020.0246} {\bibfield  {journal} {\bibinfo
  {journal} {Philos. Trans. R. Soc. A Math. Phys. Eng. Sci.}\ }\textbf
  {\bibinfo {volume} {379}},\ \bibinfo {pages} {20200246} (\bibinfo {year}
  {2021})}\BibitemShut {NoStop}%
\bibitem [{\citenamefont {Abarbanel}, \citenamefont {Rozdeba},\ and\
  \citenamefont {Shirman}(2018)}]{Abarbanel2018}%
  \BibitemOpen
  \bibfield  {author} {\bibinfo {author} {\bibfnamefont {H.~D.~I.}\
  \bibnamefont {Abarbanel}}, \bibinfo {author} {\bibfnamefont {P.~J.}\
  \bibnamefont {Rozdeba}},\ and\ \bibinfo {author} {\bibfnamefont
  {S.}~\bibnamefont {Shirman}},\ }\bibfield  {title} {\enquote {\bibinfo
  {title} {{Machine Learning: Deepest Learning as Statistical Data Assimilation
  Problems}},}\ }\href {https://doi.org/10.1162/neco_a_01094} {\bibfield
  {journal} {\bibinfo  {journal} {Neural Comput.}\ }\textbf {\bibinfo {volume}
  {30}},\ \bibinfo {pages} {2025--2055} (\bibinfo {year} {2018})}\BibitemShut
  {NoStop}%
\bibitem [{\citenamefont {Wikner}\ \emph {et~al.}(2020)\citenamefont {Wikner},
  \citenamefont {Pathak}, \citenamefont {Hunt}, \citenamefont {Girvan},
  \citenamefont {Arcomano}, \citenamefont {Szunyogh}, \citenamefont
  {Pomerance},\ and\ \citenamefont {Ott}}]{Wikner2020}%
  \BibitemOpen
  \bibfield  {author} {\bibinfo {author} {\bibfnamefont {A.}~\bibnamefont
  {Wikner}}, \bibinfo {author} {\bibfnamefont {J.}~\bibnamefont {Pathak}},
  \bibinfo {author} {\bibfnamefont {B.}~\bibnamefont {Hunt}}, \bibinfo {author}
  {\bibfnamefont {M.}~\bibnamefont {Girvan}}, \bibinfo {author} {\bibfnamefont
  {T.}~\bibnamefont {Arcomano}}, \bibinfo {author} {\bibfnamefont
  {I.}~\bibnamefont {Szunyogh}}, \bibinfo {author} {\bibfnamefont
  {A.}~\bibnamefont {Pomerance}},\ and\ \bibinfo {author} {\bibfnamefont
  {E.}~\bibnamefont {Ott}},\ }\bibfield  {title} {\enquote {\bibinfo {title}
  {Combining machine learning with knowledge-based modeling for scalable
  forecasting and subgrid-scale closure of large, complex, spatiotemporal
  systems},}\ }\href {https://doi.org/10.1063/5.0005541} {\bibfield  {journal}
  {\bibinfo  {journal} {Chaos}\ }\textbf {\bibinfo {volume} {30}},\ \bibinfo
  {pages} {053111} (\bibinfo {year} {2020})}\BibitemShut {NoStop}%
\bibitem [{\citenamefont {Fan}\ \emph {et~al.}(2020)\citenamefont {Fan},
  \citenamefont {Jiang}, \citenamefont {Zhang}, \citenamefont {Wang},\ and\
  \citenamefont {Lai}}]{Fan2020}%
  \BibitemOpen
  \bibfield  {author} {\bibinfo {author} {\bibfnamefont {H.}~\bibnamefont
  {Fan}}, \bibinfo {author} {\bibfnamefont {J.}~\bibnamefont {Jiang}}, \bibinfo
  {author} {\bibfnamefont {C.}~\bibnamefont {Zhang}}, \bibinfo {author}
  {\bibfnamefont {X.}~\bibnamefont {Wang}},\ and\ \bibinfo {author}
  {\bibfnamefont {Y.-C.}\ \bibnamefont {Lai}},\ }\bibfield  {title} {\enquote
  {\bibinfo {title} {Long-term prediction of chaotic systems with machine
  learning},}\ }\href {https://doi.org/10.1103/PhysRevResearch.2.012080}
  {\bibfield  {journal} {\bibinfo  {journal} {Phys. Rev. Research}\ }\textbf
  {\bibinfo {volume} {2}},\ \bibinfo {pages} {012080(R)} (\bibinfo {year}
  {2020})}\BibitemShut {NoStop}%
\bibitem [{\citenamefont {Wikner}\ \emph {et~al.}(2021)\citenamefont {Wikner},
  \citenamefont {Pathak}, \citenamefont {Hunt}, \citenamefont {Szunyogh},
  \citenamefont {Girvan},\ and\ \citenamefont {Ott}}]{Wikner2021}%
  \BibitemOpen
  \bibfield  {author} {\bibinfo {author} {\bibfnamefont {A.}~\bibnamefont
  {Wikner}}, \bibinfo {author} {\bibfnamefont {J.}~\bibnamefont {Pathak}},
  \bibinfo {author} {\bibfnamefont {B.~R.}\ \bibnamefont {Hunt}}, \bibinfo
  {author} {\bibfnamefont {I.}~\bibnamefont {Szunyogh}}, \bibinfo {author}
  {\bibfnamefont {M.}~\bibnamefont {Girvan}},\ and\ \bibinfo {author}
  {\bibfnamefont {E.}~\bibnamefont {Ott}},\ }\bibfield  {title} {\enquote
  {\bibinfo {title} {Using data assimilation to train a hybrid forecast system
  that combines machine-learning and knowledge-based components},}\ }\href
  {https://doi.org/10.1063/5.0048050} {\bibfield  {journal} {\bibinfo
  {journal} {Chaos}\ }\textbf {\bibinfo {volume} {31}},\ \bibinfo {pages}
  {053114} (\bibinfo {year} {2021})}\BibitemShut {NoStop}%
\bibitem [{\citenamefont {Arcomano}\ \emph {et~al.}(2022)\citenamefont
  {Arcomano}, \citenamefont {Szunyogh}, \citenamefont {Wikner}, \citenamefont
  {Pathak}, \citenamefont {Hunt},\ and\ \citenamefont {Ott}}]{Arcomano2022}%
  \BibitemOpen
  \bibfield  {author} {\bibinfo {author} {\bibfnamefont {T.}~\bibnamefont
  {Arcomano}}, \bibinfo {author} {\bibfnamefont {I.}~\bibnamefont {Szunyogh}},
  \bibinfo {author} {\bibfnamefont {A.}~\bibnamefont {Wikner}}, \bibinfo
  {author} {\bibfnamefont {J.}~\bibnamefont {Pathak}}, \bibinfo {author}
  {\bibfnamefont {B.~R.}\ \bibnamefont {Hunt}},\ and\ \bibinfo {author}
  {\bibfnamefont {E.}~\bibnamefont {Ott}},\ }\bibfield  {title} {\enquote
  {\bibinfo {title} {A hybrid approach to atmospheric modeling that combines
  machine learning with a physics-based numerical model},}\ }\href
  {https://doi.org/https://doi.org/10.1029/2021MS002712} {\bibfield  {journal}
  {\bibinfo  {journal} {Journal of Advances in Modeling Earth Systems}\
  }\textbf {\bibinfo {volume} {14}},\ \bibinfo {pages} {e2021MS002712}
  (\bibinfo {year} {2022})}\BibitemShut {NoStop}%
\bibitem [{\citenamefont {Chattopadhyay}\ \emph {et~al.}(2022)\citenamefont
  {Chattopadhyay}, \citenamefont {Mustafa}, \citenamefont {Hassanzadeh},
  \citenamefont {Bach},\ and\ \citenamefont {Kashinath}}]{Chattopadhyay2022}%
  \BibitemOpen
  \bibfield  {author} {\bibinfo {author} {\bibfnamefont {A.}~\bibnamefont
  {Chattopadhyay}}, \bibinfo {author} {\bibfnamefont {M.}~\bibnamefont
  {Mustafa}}, \bibinfo {author} {\bibfnamefont {P.}~\bibnamefont
  {Hassanzadeh}}, \bibinfo {author} {\bibfnamefont {E.}~\bibnamefont {Bach}},\
  and\ \bibinfo {author} {\bibfnamefont {K.}~\bibnamefont {Kashinath}},\
  }\bibfield  {title} {\enquote {\bibinfo {title} {Towards physics-inspired
  data-driven weather forecasting: integrating data assimilation with a deep
  spatial-transformer-based u-net in a case study with era5},}\ }\href
  {https://doi.org/10.5194/gmd-15-2221-2022} {\bibfield  {journal} {\bibinfo
  {journal} {Geoscientific Model Development}\ }\textbf {\bibinfo {volume}
  {15}},\ \bibinfo {pages} {2221--2237} (\bibinfo {year} {2022})}\BibitemShut
  {NoStop}%
\bibitem [{\citenamefont {Vlachas}\ \emph {et~al.}(2020)\citenamefont
  {Vlachas}, \citenamefont {Pathak}, \citenamefont {Hunt}, \citenamefont
  {Sapsis}, \citenamefont {Girvan}, \citenamefont {Ott},\ and\ \citenamefont
  {Koumoutsakos}}]{vlachas2019}%
  \BibitemOpen
  \bibfield  {author} {\bibinfo {author} {\bibfnamefont {P.}~\bibnamefont
  {Vlachas}}, \bibinfo {author} {\bibfnamefont {J.}~\bibnamefont {Pathak}},
  \bibinfo {author} {\bibfnamefont {B.}~\bibnamefont {Hunt}}, \bibinfo {author}
  {\bibfnamefont {T.}~\bibnamefont {Sapsis}}, \bibinfo {author} {\bibfnamefont
  {M.}~\bibnamefont {Girvan}}, \bibinfo {author} {\bibfnamefont
  {E.}~\bibnamefont {Ott}},\ and\ \bibinfo {author} {\bibfnamefont
  {P.}~\bibnamefont {Koumoutsakos}},\ }\bibfield  {title} {\enquote {\bibinfo
  {title} {Backpropagation algorithms and reservoir computing in recurrent
  neural networks for the forecasting of complex spatiotemporal dynamics},}\
  }\href {https://doi.org/https://doi.org/10.1016/j.neunet.2020.02.016}
  {\bibfield  {journal} {\bibinfo  {journal} {Neural Netw.}\ }\textbf {\bibinfo
  {volume} {126}},\ \bibinfo {pages} {191 -- 217} (\bibinfo {year}
  {2020})}\BibitemShut {NoStop}%
\bibitem [{\citenamefont {Parlitz}\ and\ \citenamefont
  {Merkwirth}(2000)}]{Parlitz2000}%
  \BibitemOpen
  \bibfield  {author} {\bibinfo {author} {\bibfnamefont {U.}~\bibnamefont
  {Parlitz}}\ and\ \bibinfo {author} {\bibfnamefont {C.}~\bibnamefont
  {Merkwirth}},\ }\bibfield  {title} {\enquote {\bibinfo {title} {Prediction of
  spatiotemporal time series based on reconstructed local states},}\ }\href
  {https://doi.org/10.1103/PhysRevLett.84.1890} {\bibfield  {journal} {\bibinfo
   {journal} {Phys. Rev. Lett.}\ }\textbf {\bibinfo {volume} {84}},\ \bibinfo
  {pages} {1890--1893} (\bibinfo {year} {2000})}\BibitemShut {NoStop}%
\bibitem [{\citenamefont {Ørstavik}\ and\ \citenamefont
  {Stark}(1998)}]{Orstavik1998}%
  \BibitemOpen
  \bibfield  {author} {\bibinfo {author} {\bibfnamefont {S.}~\bibnamefont
  {Ørstavik}}\ and\ \bibinfo {author} {\bibfnamefont {J.}~\bibnamefont
  {Stark}},\ }\bibfield  {title} {\enquote {\bibinfo {title} {Reconstruction
  and cross-prediction in coupled map lattices using spatio-temporal embedding
  techniques},}\ }\href
  {https://doi.org/https://doi.org/10.1016/S0375-9601(98)00541-6} {\bibfield
  {journal} {\bibinfo  {journal} {Physics Letters A}\ }\textbf {\bibinfo
  {volume} {247}},\ \bibinfo {pages} {145--160} (\bibinfo {year}
  {1998})}\BibitemShut {NoStop}%
\bibitem [{\citenamefont {Gauthier}\ \emph {et~al.}(2021)\citenamefont
  {Gauthier}, \citenamefont {Bollt}, \citenamefont {Griffith},\ and\
  \citenamefont {Barbosa}}]{NG-RC}%
  \BibitemOpen
  \bibfield  {author} {\bibinfo {author} {\bibfnamefont {D.~J.}\ \bibnamefont
  {Gauthier}}, \bibinfo {author} {\bibfnamefont {E.}~\bibnamefont {Bollt}},
  \bibinfo {author} {\bibfnamefont {A.}~\bibnamefont {Griffith}},\ and\
  \bibinfo {author} {\bibfnamefont {W.~A.~S.}\ \bibnamefont {Barbosa}},\
  }\bibfield  {title} {\enquote {\bibinfo {title} {Next generation reservoir
  computing},}\ }\href {https://doi.org/10.1038/s41467-021-25801-2} {\bibfield
  {journal} {\bibinfo  {journal} {Nat. Commun.}\ }\textbf {\bibinfo {volume}
  {12}},\ \bibinfo {pages} {5564} (\bibinfo {year} {2021})}\BibitemShut
  {NoStop}%
\bibitem [{\citenamefont {Lai}(2021)}]{Lai2021}%
  \BibitemOpen
  \bibfield  {author} {\bibinfo {author} {\bibfnamefont {Y.-C.}\ \bibnamefont
  {Lai}},\ }\bibfield  {title} {\enquote {\bibinfo {title} {Finding nonlinear
  system equations and complex network structures from data: A sparse
  optimization approach},}\ }\href {https://doi.org/10.1063/5.0062042}
  {\bibfield  {journal} {\bibinfo  {journal} {Chaos}\ }\textbf {\bibinfo
  {volume} {31}},\ \bibinfo {pages} {082101} (\bibinfo {year}
  {2021})}\BibitemShut {NoStop}%
\bibitem [{\citenamefont {Bollt}(2021)}]{Bollt}%
  \BibitemOpen
  \bibfield  {author} {\bibinfo {author} {\bibfnamefont {E.}~\bibnamefont
  {Bollt}},\ }\bibfield  {title} {\enquote {\bibinfo {title} {On explaining the
  surprising success of reservoir computing forecaster of chaos? the universal
  machine learning dynamical system with contrast to var and dmd},}\ }\href
  {https://doi.org/10.1063/5.0024890} {\bibfield  {journal} {\bibinfo
  {journal} {Chaos}\ }\textbf {\bibinfo {volume} {31}},\ \bibinfo {pages}
  {013108} (\bibinfo {year} {2021})}\BibitemShut {NoStop}%
\bibitem [{\citenamefont {Pathak}\ \emph
  {et~al.}(2018{\natexlab{a}})\citenamefont {Pathak}, \citenamefont {Wikner},
  \citenamefont {Fussell}, \citenamefont {Chandra}, \citenamefont {Hunt},
  \citenamefont {Girvan},\ and\ \citenamefont {Ott}}]{Pathak2018Hybrid}%
  \BibitemOpen
  \bibfield  {author} {\bibinfo {author} {\bibfnamefont {J.}~\bibnamefont
  {Pathak}}, \bibinfo {author} {\bibfnamefont {A.}~\bibnamefont {Wikner}},
  \bibinfo {author} {\bibfnamefont {R.}~\bibnamefont {Fussell}}, \bibinfo
  {author} {\bibfnamefont {S.}~\bibnamefont {Chandra}}, \bibinfo {author}
  {\bibfnamefont {B.~R.}\ \bibnamefont {Hunt}}, \bibinfo {author}
  {\bibfnamefont {M.}~\bibnamefont {Girvan}},\ and\ \bibinfo {author}
  {\bibfnamefont {E.}~\bibnamefont {Ott}},\ }\bibfield  {title} {\enquote
  {\bibinfo {title} {Hybrid forecasting of chaotic processes: Using machine
  learning in conjunction with a knowledge-based model},}\ }\href
  {https://doi.org/10.1063/1.5028373} {\bibfield  {journal} {\bibinfo
  {journal} {Chaos}\ }\textbf {\bibinfo {volume} {28}},\ \bibinfo {pages}
  {041101} (\bibinfo {year} {2018}{\natexlab{a}})}\BibitemShut {NoStop}%
\bibitem [{\citenamefont {Pathak}\ \emph
  {et~al.}(2018{\natexlab{b}})\citenamefont {Pathak}, \citenamefont {Hunt},
  \citenamefont {Girvan}, \citenamefont {Lu},\ and\ \citenamefont
  {Ott}}]{Pathak2018}%
  \BibitemOpen
  \bibfield  {author} {\bibinfo {author} {\bibfnamefont {J.}~\bibnamefont
  {Pathak}}, \bibinfo {author} {\bibfnamefont {B.}~\bibnamefont {Hunt}},
  \bibinfo {author} {\bibfnamefont {M.}~\bibnamefont {Girvan}}, \bibinfo
  {author} {\bibfnamefont {Z.}~\bibnamefont {Lu}},\ and\ \bibinfo {author}
  {\bibfnamefont {E.}~\bibnamefont {Ott}},\ }\bibfield  {title} {\enquote
  {\bibinfo {title} {Model-free prediction of large spatiotemporally chaotic
  systems from data: A reservoir computing approach},}\ }\href
  {https://doi.org/10.1103/PhysRevLett.120.024102} {\bibfield  {journal}
  {\bibinfo  {journal} {Phys. Rev. Lett.}\ }\textbf {\bibinfo {volume} {120}},\
  \bibinfo {pages} {024102} (\bibinfo {year} {2018}{\natexlab{b}})}\BibitemShut
  {NoStop}%
\bibitem [{\citenamefont {Lu}\ \emph {et~al.}(2017)\citenamefont {Lu},
  \citenamefont {Pathak}, \citenamefont {Hunt}, \citenamefont {Girvan},
  \citenamefont {Brockett},\ and\ \citenamefont {Ott}}]{Lu2017}%
  \BibitemOpen
  \bibfield  {author} {\bibinfo {author} {\bibfnamefont {Z.}~\bibnamefont
  {Lu}}, \bibinfo {author} {\bibfnamefont {J.}~\bibnamefont {Pathak}}, \bibinfo
  {author} {\bibfnamefont {B.}~\bibnamefont {Hunt}}, \bibinfo {author}
  {\bibfnamefont {M.}~\bibnamefont {Girvan}}, \bibinfo {author} {\bibfnamefont
  {R.}~\bibnamefont {Brockett}},\ and\ \bibinfo {author} {\bibfnamefont
  {E.}~\bibnamefont {Ott}},\ }\bibfield  {title} {\enquote {\bibinfo {title}
  {Reservoir observers: Model-free inference of unmeasured variables in chaotic
  systems},}\ }\href {https://doi.org/10.1063/1.4979665} {\bibfield  {journal}
  {\bibinfo  {journal} {Chaos}\ }\textbf {\bibinfo {volume} {27}},\ \bibinfo
  {pages} {041102} (\bibinfo {year} {2017})}\BibitemShut {NoStop}%
\bibitem [{\citenamefont {Herteux}\ and\ \citenamefont
  {R{\"a}th}(2020)}]{Herteux2020}%
  \BibitemOpen
  \bibfield  {author} {\bibinfo {author} {\bibfnamefont {J.}~\bibnamefont
  {Herteux}}\ and\ \bibinfo {author} {\bibfnamefont {C.}~\bibnamefont
  {R{\"a}th}},\ }\bibfield  {title} {\enquote {\bibinfo {title} {Breaking
  symmetries of the reservoir equations in echo state networks},}\ }\href
  {https://doi.org/10.1063/5.0028993} {\bibfield  {journal} {\bibinfo
  {journal} {Chaos}\ }\textbf {\bibinfo {volume} {30}},\ \bibinfo {pages}
  {123142} (\bibinfo {year} {2020})}\BibitemShut {NoStop}%
\bibitem [{\citenamefont {Barbosa}\ \emph {et~al.}(2021)\citenamefont
  {Barbosa}, \citenamefont {Griffith}, \citenamefont {Rowlands}, \citenamefont
  {Govia}, \citenamefont {Ribeill}, \citenamefont {Nguyen}, \citenamefont
  {Ohki},\ and\ \citenamefont {Gauthier}}]{BarbosaPRE2021}%
  \BibitemOpen
  \bibfield  {author} {\bibinfo {author} {\bibfnamefont {W.~A.~S.}\
  \bibnamefont {Barbosa}}, \bibinfo {author} {\bibfnamefont {A.}~\bibnamefont
  {Griffith}}, \bibinfo {author} {\bibfnamefont {G.~E.}\ \bibnamefont
  {Rowlands}}, \bibinfo {author} {\bibfnamefont {L.~C.~G.}\ \bibnamefont
  {Govia}}, \bibinfo {author} {\bibfnamefont {G.~J.}\ \bibnamefont {Ribeill}},
  \bibinfo {author} {\bibfnamefont {M.-H.}\ \bibnamefont {Nguyen}}, \bibinfo
  {author} {\bibfnamefont {T.~A.}\ \bibnamefont {Ohki}},\ and\ \bibinfo
  {author} {\bibfnamefont {D.~J.}\ \bibnamefont {Gauthier}},\ }\bibfield
  {title} {\enquote {\bibinfo {title} {Symmetry-aware reservoir computing},}\
  }\href {https://doi.org/10.1103/PhysRevE.104.045307} {\bibfield  {journal}
  {\bibinfo  {journal} {Phys. Rev. E}\ }\textbf {\bibinfo {volume} {104}},\
  \bibinfo {pages} {045307} (\bibinfo {year} {2021})}\BibitemShut {NoStop}%
\bibitem [{\citenamefont {Favoni}\ \emph {et~al.}(2022)\citenamefont {Favoni},
  \citenamefont {Ipp}, \citenamefont {M\"uller},\ and\ \citenamefont
  {Schuh}}]{Favoni2022}%
  \BibitemOpen
  \bibfield  {author} {\bibinfo {author} {\bibfnamefont {M.}~\bibnamefont
  {Favoni}}, \bibinfo {author} {\bibfnamefont {A.}~\bibnamefont {Ipp}},
  \bibinfo {author} {\bibfnamefont {D.~I.}\ \bibnamefont {M\"uller}},\ and\
  \bibinfo {author} {\bibfnamefont {D.}~\bibnamefont {Schuh}},\ }\bibfield
  {title} {\enquote {\bibinfo {title} {Lattice gauge equivariant convolutional
  neural networks},}\ }\href {https://doi.org/10.1103/PhysRevLett.128.032003}
  {\bibfield  {journal} {\bibinfo  {journal} {Phys. Rev. Lett.}\ }\textbf
  {\bibinfo {volume} {128}},\ \bibinfo {pages} {032003} (\bibinfo {year}
  {2022})}\BibitemShut {NoStop}%
\bibitem [{\citenamefont {Oberlack}\ \emph {et~al.}(2022)\citenamefont
  {Oberlack}, \citenamefont {Hoyas}, \citenamefont {Kraheberger}, \citenamefont
  {Alc\'antara-\'Avila},\ and\ \citenamefont {Laux}}]{OberlackPRL2022}%
  \BibitemOpen
  \bibfield  {author} {\bibinfo {author} {\bibfnamefont {M.}~\bibnamefont
  {Oberlack}}, \bibinfo {author} {\bibfnamefont {S.}~\bibnamefont {Hoyas}},
  \bibinfo {author} {\bibfnamefont {S.~V.}\ \bibnamefont {Kraheberger}},
  \bibinfo {author} {\bibfnamefont {F.}~\bibnamefont {Alc\'antara-\'Avila}},\
  and\ \bibinfo {author} {\bibfnamefont {J.}~\bibnamefont {Laux}},\ }\bibfield
  {title} {\enquote {\bibinfo {title} {Turbulence statistics of arbitrary
  moments of wall-bounded shear flows: A symmetry approach},}\ }\href
  {https://doi.org/10.1103/PhysRevLett.128.024502} {\bibfield  {journal}
  {\bibinfo  {journal} {Phys. Rev. Lett.}\ }\textbf {\bibinfo {volume} {128}},\
  \bibinfo {pages} {024502} (\bibinfo {year} {2022})}\BibitemShut {NoStop}%
\bibitem [{\citenamefont {Liu}\ and\ \citenamefont {Tegmark}(2022)}]{Liu2022}%
  \BibitemOpen
  \bibfield  {author} {\bibinfo {author} {\bibfnamefont {Z.}~\bibnamefont
  {Liu}}\ and\ \bibinfo {author} {\bibfnamefont {M.}~\bibnamefont {Tegmark}},\
  }\bibfield  {title} {\enquote {\bibinfo {title} {Machine learning hidden
  symmetries},}\ }\href {https://doi.org/10.1103/PhysRevLett.128.180201}
  {\bibfield  {journal} {\bibinfo  {journal} {Phys. Rev. Lett.}\ }\textbf
  {\bibinfo {volume} {128}},\ \bibinfo {pages} {180201} (\bibinfo {year}
  {2022})}\BibitemShut {NoStop}%
\bibitem [{\citenamefont {Lorenz}(2005)}]{Lorenz2005}%
  \BibitemOpen
  \bibfield  {author} {\bibinfo {author} {\bibfnamefont {E.~N.}\ \bibnamefont
  {Lorenz}},\ }\bibfield  {title} {\enquote {\bibinfo {title} {Designing
  chaotic models},}\ }\href {https://doi.org/https://doi.org/10.1175/JAS3430.1}
  {\bibfield  {journal} {\bibinfo  {journal} {J. Atmos. Sci.}\ }\textbf
  {\bibinfo {volume} {62}},\ \bibinfo {pages} {1574--1587} (\bibinfo {year}
  {2005})}\BibitemShut {NoStop}%
\bibitem [{\citenamefont {Thornes}, \citenamefont {D{\"u}ben},\ and\
  \citenamefont {Palmer}(2017)}]{Thornes2017}%
  \BibitemOpen
  \bibfield  {author} {\bibinfo {author} {\bibfnamefont {T.}~\bibnamefont
  {Thornes}}, \bibinfo {author} {\bibfnamefont {P.}~\bibnamefont {D{\"u}ben}},\
  and\ \bibinfo {author} {\bibfnamefont {T.}~\bibnamefont {Palmer}},\
  }\bibfield  {title} {\enquote {\bibinfo {title} {On the use of
  scale-dependent precision in earth system modelling},}\ }\href
  {https://doi.org/https://doi.org/10.1002/qj.2974} {\bibfield  {journal}
  {\bibinfo  {journal} {Q. J. R. Meteorol. Soc.}\ }\textbf {\bibinfo {volume}
  {143}},\ \bibinfo {pages} {897--908} (\bibinfo {year} {2017})}\BibitemShut
  {NoStop}%
\bibitem [{\citenamefont {Chattopadhyay}, \citenamefont {Subel},\ and\
  \citenamefont {Hassanzadeh}(2020)}]{ChattopadhyayTransfLearning2020}%
  \BibitemOpen
  \bibfield  {author} {\bibinfo {author} {\bibfnamefont {A.}~\bibnamefont
  {Chattopadhyay}}, \bibinfo {author} {\bibfnamefont {A.}~\bibnamefont
  {Subel}},\ and\ \bibinfo {author} {\bibfnamefont {P.}~\bibnamefont
  {Hassanzadeh}},\ }\bibfield  {title} {\enquote {\bibinfo {title} {Data-driven
  super-parameterization using deep learning: Experimentation with multiscale
  lorenz 96 systems and transfer learning},}\ }\href
  {https://doi.org/https://doi.org/10.1029/2020MS002084} {\bibfield  {journal}
  {\bibinfo  {journal} {J. Adv. Model. Earth Syst.}\ }\textbf {\bibinfo
  {volume} {12}},\ \bibinfo {pages} {e2020MS002084} (\bibinfo {year}
  {2020})}\BibitemShut {NoStop}%
\bibitem [{Ing()}]{IngoPrivate}%
  \BibitemOpen
  \href@noop {} {}\bibinfo {note} {Mirko Goldmann, Claudio R. Mirasso, Ingo
  Fischer and Miguel C. Soriano, private communication.}\BibitemShut {Stop}%
\bibitem [{\citenamefont {Platt}\ \emph {et~al.}()\citenamefont {Platt},
  \citenamefont {Penny}, \citenamefont {Smith}, \citenamefont {Chen},\ and\
  \citenamefont {Abarbanel}}]{Platt2022}%
  \BibitemOpen
  \bibfield  {author} {\bibinfo {author} {\bibfnamefont {J.~A.}\ \bibnamefont
  {Platt}}, \bibinfo {author} {\bibfnamefont {S.~G.}\ \bibnamefont {Penny}},
  \bibinfo {author} {\bibfnamefont {T.~A.}\ \bibnamefont {Smith}}, \bibinfo
  {author} {\bibfnamefont {T.-C.}\ \bibnamefont {Chen}},\ and\ \bibinfo
  {author} {\bibfnamefont {H.~D.~I.}\ \bibnamefont {Abarbanel}},\ }\href@noop
  {} {}\Eprint {https://arxiv.org/abs/2201.08910} {arXiv:2201.08910}
  \BibitemShut {NoStop}%
\end{thebibliography}

%

\end{document}